\documentclass{acmsmall} 
\usepackage{tikz}
\usetikzlibrary{arrows}

\definecolor{mahogany}{rgb}{0.75, 0.25, 0.0}
\definecolor{navyblue}{rgb}{0.0, 0.0, 0.5}
\definecolor{onyx}{rgb}{0.06, 0.06, 0.06}
\definecolor{richblack}{rgb}{0.0, 0.25, 0.25}
\definecolor{rosewood}{rgb}{0.4, 0.0, 0.04}
\definecolor{maroonh}{rgb}{0.5, 0.0, 0.0}
\definecolor{maroonx}{rgb}{0.69, 0.19, 0.38}
\definecolor{upmaroon}{rgb}{0.48, 0.07, 0.07}

\usepackage{mathtools,dsfont,amsthm,amsfonts,amsmath,cite,amssymb,datetime,enumerate}
\mathtoolsset{
showonlyrefs=true
}
\usepackage[colorlinks,pdfpagelabels,linkcolor=onyx,citecolor=blue,plainpages=false]{hyperref}

        {\hspace*{\fill}$\Box$\par\vspace{4mm}}
\newenvironment{guarantee}{\vspace{4mm}\noindent{\emph{Guarantee.}}}%
        {\hspace*{\fill}$\Box$\par\vspace{4mm}}

\newcommand*\btheta{{\ensuremath{\boldsymbol{\theta}}}}
\newcommand*\bphi{{\ensuremath{\boldsymbol{\phi}}}}
\newcommand*\bmu{\ensuremath{\boldsymbol\mu}}
\newcommand*\bepsilon{\ensuremath{\boldsymbol\epsilon}}

\DeclareMathOperator*{\argmax}{\mathtt{argmax}}
\DeclareMathOperator*{\argmin}{\mathtt{argmin}}
\DeclareMathOperator*{\argopt}{\mathtt{argopt}}
\DeclareMathOperator*{\arglocmin}{\mathtt{arglocmin}}
\DeclareMathOperator*{\arglocmax}{\mathtt{arglocmax}}
\DeclareMathOperator*{\arglocopt}{\mathtt{arglocopt}}

\newcommand*\loss{\ensuremath{\boldsymbol\ell}}
\DeclareMathOperator*{\expec}{\mathbb E}
\newcommand{\grad}{\operatorname{\nabla}}
\newcommand{\quer}{\mathsf{Q}}
\newcommand{\resp}{\mathsf{R}}

\newcommand{\eod}{{${}$\\}}

\newcommand{\sta}{\mathbf{s}}
\newcommand{\act}{\mathbf{a}}
\newcommand{\bI}{\mathbf{I}}

\newcommand{\fobj}{{\mathcal R}}

\newcommand{\cF}{{\mathcal F}}

\newcommand{\cS}{{\mathcal S}}
\newcommand{\cP}{{\mathcal P}}

\newcommand{\cA}{{\mathcal A}}
\newcommand{\vlb}{{\mathcal L}}

\newcommand{\bR}{{\mathbb R}}

\newcommand{\bP}{{\mathbb P}}
\newcommand{\bbO}{{\mathds 1}}
\newcommand{\bQ}{{\mathbb Q}}

\newcommand{\encoder}{{\mathbb E}}
\newcommand{\decoder}{{\mathbb D}}

\newcommand{\vt}{{\mathbf v}}
\newcommand{\Wt}{{\mathbf W}}
\newcommand{\Vt}{{\mathbf V}}
\newcommand{\Gd}{{\mathbf G}}
\newcommand{\x}{{\mathbf x}}

\newcommand{\z}{{\mathbf z}}

\newtheorem{defn}{Definition}

\newtheorem{rem}{Remark}
\newtheorem{coupling}{Grammar}
\newtheorem{repn}{Representation}

\tikzstyle{no_block} = [circle,minimum size=2em]
\tikzstyle{wblock} = [draw,fill=yellow!20,minimum size=1em]
\tikzstyle{block} = [draw,fill=black!20,minimum size=1em]
\tikzstyle{o_block} = [draw,fill=black!20,minimum size=1em]
\tikzstyle{red_block} = [draw,fill=blue!20,minimum size=1em]
\tikzstyle{green_block} = [draw,fill=black!100,minimum size=1em]
\tikzstyle{yellow_block} = [draw,fill=black!100,minimum size=1em]
 
\tikzstyle{branch}=[fill,shape=circle,minimum size=3pt,inner sep=0pt]

\begin{document}

\markboth{Balduzzi}{Semantics, Representations and Grammars for Deep Learning}
\title{Semantics, Representations and Grammars for Deep Learning}
\author{David Balduzzi}

\begin{abstract}
	Deep learning is currently the subject of intensive study. However, fundamental concepts such as representations are not formally defined -- researchers ``know them when they see them'' -- and there is no common language for describing and analyzing algorithms. This essay proposes an abstract framework that identifies the essential features of current practice and may provide a foundation for future developments. 

 	The backbone of almost all deep learning algorithms is backpropagation, which is simply a gradient computation distributed over a neural network. The main ingredients of the framework are thus, unsurprisingly: (i) game theory, to formalize distributed optimization; and (ii) communication protocols, to track the flow of zeroth and first-order information. The framework allows natural definitions of semantics (as the meaning encoded in functions), representations (as functions whose semantics is chosen to optimized a criterion) and grammars (as communication protocols equipped with first-order convergence guarantees).
 	
 	Much of the essay is spent discussing examples taken from the literature. The ultimate aim is to develop a graphical language for describing the structure of deep learning algorithms that backgrounds the details of the optimization procedure and foregrounds how the components interact.  Inspiration is taken from probabilistic graphical models and factor graphs, which capture the essential structural features of multivariate distributions.
\end{abstract}

\maketitle

\terms{deep learning; representation learning; optimization; game theory; neural networks}

\setcounter{tocdepth}{2}{\sffamily{\tableofcontents}}\vspace{1cm}

\setcounter{section}{-1}

\section{Introduction}

Deep learning has achieved remarkable successes in object and voice recognition, machine translation, reinforcement learning and other tasks \cite{krizhevsky:12, hinton:12b, sutskever:14, Mnih:2015wq, lecun:15}. From a practical standpoint the problem of supervised learning is well-understood and has largely been solved  -- at least in the regime where both labeled data and computational power are abundant. The workhorse underlying most deep learning algorithms is error backpropagation \cite{werbos:74,rumelhart:86,rumelhart:86a,schmidhuber:15}, which is simply gradient descent distributed across a neural network via the chain rule. 

Gradient descent and its variants are well-understood when applied to convex or nearly convex objectives \cite{robbins:51,nemirovski:78,nemirovski:79,nemirovski:09}. In particular, they have strong performance guarantees in the stochastic and adversarial settings \cite{zinkevich:03,cesa:06,bottou:10,shalev:11}. The reasons for the success of gradient descent in non-convex settings are less clear, although recent work has provided evidence that most local minima are good enough \cite{choromanska:14,choromanska:15}; that modern convolutional networks are close enough to convex for many results on rates of convergence apply \cite{doco:15}; and that the rate of convergence of gradient-descent can control generalization performance, even in nonconvex settings \cite{hardt:15}.

Taking a step back, gradient-based optimization provides a well-established set of computational primitives \cite{gordon:06}, with theoretical backing in simple cases and empirical backing in others. First-order optimization thus falls in broadly the same category as computing an eigenvector or inverting a matrix: given sufficient data and computational resources, we have algorithms that reliably find good enough solutions for a wide range of problems. 

This essay proposes to abstract out the optimization algorithms used for weight updates and focus on how the components of deep learning algorithms interact.  Treating optimization as a computational primitive encourages a shift from low-level algorithm design to higher-level mechanism design: we can shift attention to designing architectures that are guaranteed to learn distributed representations suited to specific objectives. The goal is to introduce a language at a level of abstraction where designers can focus on formal specifications (grammars) that specify how plug-and-play optimization modules combine into larger learning systems.

\subsection{What is a representation?}

Let us recall how representation learning is commonly understood. Bengio \emph{et al} describe representation learning as ``learning transformations of the data that make it easier to extract useful information when building classifiers or other predictors'' \cite{bengio:13}. More specifically, ``a deep learning algorithm is a particular kind of representation learning procedure that discovers multiple levels of representation, with higher-level features representing more abstract aspects of the data'' \cite{bengio:13a}. Finally, LeCun \emph{et al} state that multiple levels of representations are obtained ``by composing simple but non-linear modules that each transform the representation at one level (starting with the raw input) into a representation at a higher, slightly more abstract level. With the composition of enough such transformations, very complex functions can be learned. For classification tasks, higher layers of representation amplify aspects of the input that are important for discrimination and suppress irrelevant variations'' \cite{lecun:15}.

The quotes describe the operation of a successful deep learning algorithm. What is lacking is a characterization of what makes a deep learning algorithm work in the first place. What properties must an algorithm have to learn layered representations? What does it mean for the representation learned by one layer to be useful to another? What, exactly, is a representation?

In practice, almost all deep learning algorithms rely on error backpropagation to ``align'' the representations learned by different layers of a network. This suggests that the answers to the above questions are tightly bound up in first-order (that is, gradient-based) optimization methods. It is therefore unsurprisingly that the bulk of the paper is concerned with tracking the flow of first-order information. The framework is intended to facilitate the design of more general first-order algorithms than backpropagation.

\vspace{2mm}
\noindent
\textbf{\sffamily{Semantics.}}
To get started, we need a theory of the meaning or semantics encoded in neural networks. Since there is nothing special about neural networks, the approach taken is inclusive and minimalistic. Definition~\ref{d:meaning} states that the meaning of \emph{any} function is how it implicitly categorizes inputs by assigning them to outputs. The next step is to characterize those functions whose semantics encode knowledge, and for this we turn to optimization \cite{sra:12}. 

\vspace{2mm}
\noindent
\textbf{\sffamily{Representations from optimizations.}}
Nemirovski and Yudin developed the black-box computational model to analyze the computational complexity of first-order optimization methods \cite{nemirovski:83,agarwal:09,raginsky:11,arjevani:15}. The black-box model is a more abstract view on optimization than the Turing machine model: it specifies a \emph{communication protocol} that tracks how often an algorithm makes \emph{queries} about the objective. It is useful to refine Nemirovski and Yudin's terminology by distinguishing between black-boxes, which \emph{respond} with zeroth-order information (the value of a function at the query-point), and gray-boxes\footnote{Gray for gradient.}, which respond with zeroth- and first-order information (the gradient or subgradient). 

With these preliminaries in hand, Definition~\ref{d:foo} proposes that a \emph{representation} is a function that is a \emph{local} solution to an optimization problem. Since we do not restrict to convex problems, finding global solutions is not feasible. Indeed, recent experience shows that global solutions are often not necessary practice \cite{krizhevsky:12, hinton:12b, sutskever:14, Mnih:2015wq, lecun:15}. The local solution has similar semantics to -- that is, it represents -- the ideal solution. The ideal solution usually cannot be found: due to computational limitations, since the problem is nonconvex, because we only have access to a finite sample from an unknown distribution, etc.

To see how Definition~\ref{d:foo} connects with representation learning as commonly understood, it is necessary to take a detour through distributed optimization and game theory.

\subsection{Distributed representations}

Game theory provides tools for analyzing distributed optimization problems where a set of players aim to minimizes losses that depend not only on their actions, but also the actions of all other players in the game \cite{vonneumann:44,agt:07}. Game theory has traditionally focused on convex losses since they are more theoretically amenable. Here, the only restriction imposed on losses is that they are differentiable almost everywhere. 

Allowing nonconvex losses means that error-backpropagation can be reformulated as a game. Interestingly, there is enormous freedom in choosing the players. They can correspond to individual units, layers, entire neural networks, and a variety of other, intermediate choices. An advantage of the game-theoretic formulation is thus that it applies at many different scales. 

Nonconvex losses and local optima are essential to developing a \emph{scale-free} formalism. Even when it turns out that particular units or a particular layer of a neural network are solving a convex problem, convexity is destroyed as soon as those units or layers are combined to form larger learning systems. Convexity is not a property that is preserved in general when units are combined into layers or layers into networks. It is therefore convenient to introduce the computational primitive $\arglocopt$ to denote the output of a first-order optimization procedure, see Definition~\ref{d:foo}.

\vspace{2mm}
\noindent
\textbf{\sffamily{A concern about excessive generality.}}
A potential criticism is that the formulation is too broad. Very little can be said about nonconvex optimization in general; introducing games where many players jointly optimize a set of arbitary nonconvex functions only compounds the problem. 

Additional structure is required. A successful case study can be found in \cite{doco:15}, which presents a detailed game-theoretic analysis of rectifier neural networks. The key to the analysis is that rectifier units are almost convex. The main result is that the rate of convergence of a neural network to a local optimum is controlled by the (waking-)regret of the algorithms applied to compute weight updates in the network. 

Whereas \cite{doco:15} relied heavily on specific properties of rectifer nonlinearities, this paper considers a wide-range of deep learning architectures. Nevertheless, it is possible to carve out an interesting subclass of nonconvex games by identifying the composition of simple functions as an essential feature common to deep learning architectures. Compositionality is formalized via distributed communication protocols and grammars.

\vspace{2mm}
\noindent
\textbf{\sffamily{Grammars for games.}}
Neural networks are constructed by composing a series of elementary operations. The resulting feedforward computation is captured via as a computation graph \cite{griewank:08,baydin:14,bergstra:10,bastien:12,merrienboer:15,schulman:15}. Backpropagation traverses the graph in reverse and recursively computes the gradient with respect to the parameters at each node.

Section~\ref{sec:comp} maps the feedforward and feedback computations onto the queries and responses that arise in Nemirovski and Yudin's model of optimization. However, queries and responses are now highly structured. In the query phase, players feed parameters into a computation graph (the Query graph $\quer$) that performs the feedforward sweep. In the response phase, oracles reveal first-order information that is fed into a second computation graph (the Response graph $\resp$). 

In most cases the Response graph simply implements backpropagation. However, there are examples where it does not. Three are highlighted here, see section~\ref{sec:advers}, and especially sections~\ref{sec:pg} and \ref{sec:kb}. Other algorithms where the Response graphs do not simply implement backprop include difference target propagation \cite{lee:15} and feedback alignment \cite{lillicrap:14} (both discussed briefly in section~\ref{sec:kb}) and truncated backpropagation through time \cite{elman:90,williams:90,williams:95}, where a choice is made about where to cut backprop short. Examples where the query and response graph differ are of particular interest, since they point towards more general classes of deep learning algorithms.

A \emph{distributed communication protocol} is a game with additional structure: the Query and Response graphs, see Definition~\ref{d:dcp}. The graphs capture the compositional structure of the functions learned by a neural network and the compositional structure of the learning procedure respectively. It is important for our purposes that (i) the feedforward and feedback sweeps correspond to two distinct graphs and (ii) the communication protocol is kept distinct from the optimization procedure. That is, the communication protocol specifies how information flows through the networks without specifying how players make use of it. Players can be treated as plug-and-play rational agents that are provided with carefully constructed and coordinated first-order information to optimize as they see fit \cite{Russel:09,gershman:15}. 

Finally, a \emph{grammar} is a distributed communication protocol equipped with a guarantee that the response graph encodes sufficient information for the players to jointly find a local optimum of an objective function. The paradigmatic example of a grammar is backpropagation. A grammar is a thus a game designed to perform a task. A representation learned by one (p)layer is useful to another if the game is guaranteed to converge on a local solution to an objective -- that is, if the players interact though a grammar. It follows that the players build representations that jointly encode knowledge about the task.

\vspace{2mm}
\noindent
\textbf{\sffamily{Caveats.}}
What follows is provisional. The definitions are a first attempt to capture an interesting, and perhaps useful, perspective on deep learning. The essay contains no new theorems, algorithms or experiments, see \cite{bvb:15,doco:15,gradprop:15} for ``real work'' based on the ideas presented here. The essay is not intended to be comprehensive. Many details are left out and many important aspects are not covered: most notably, probabilistic and Bayesian formulations, and various methods for unsupervised pre-training.

\vspace{2mm}
\noindent
\textbf{\sffamily{A series of worked examples.}}
In line with its provisional nature, much of the essay is spent applying the framework to worked examples: error backpropagation as a supervised model \cite{rumelhart:86a}; variational autoencoders \cite{kingma:14} and generative adversarial networks \cite{goodfellow:14} for unsupervised learning; the deviator-actor-critic (DAC) model for deep reinforcement learning \cite{gradprop:15}; and kickback, a biologically plausible variant of backpropagation \cite{bvb:15}. The examples were chosen, in part, to maximize variety and, in part, based on familiarity. The discussions are short; the interested reader is encouraged to consult the original papers to fill in the gaps.

The last two examples are particularly interesting since their Response graphs differ substantially from backpropagation. The DAC model constructs a zeroth-order black-box to estimate gradients rather than querying a first-order gray-box. Kickback prunes backprop's Response graph by replacing most of its gray-boxes with black-boxes and approximating the chain rule with (primarily) local computations.

\subsection{Related work}

Bottou and Gallinari proposed to decompose neural networks into cooperating modules \cite{bottou:91,bottou:14}. Decomposing more general algorithms or models into collections of interacting agents dates back to the shrieking demons that comprised Selfridge's Pandemonium \cite{selfridge:58} and a long line of related work \cite{klopf:82,barto:85,minsky:86,baum:99,kwee:01,vonbartheld:01,seung:03,lewis:14}. The focus on components of neural networks as players, or rational agents, in their own right developed here derives from work aimed at modeling biological neurons game-theoretically, see \cite{bb:12,bob:13,bt:13,balduzzi:13mv,balduzzi:14cpm}.

A related approach to semantics based on general value functions can be found in Sutton \emph{et al} \cite{sutton:11}, see remark~\ref{rem:sutton}. Computation graphs as applied to backprop are the basis of the Python library Theano \cite{bergstra:10,bastien:12,merrienboer:15} and provide the backbone for automatic/algorithmic differentiation \cite{griewank:08,baydin:14}. 

Grammars are a technical term in the theory of formal languages relating to the Chomsky hierarchy \cite{hopcroft:79}. There is no apparent relation between that notion of grammar and the one presented here, aside from both relating to structural rules governing composition. Formal languages and deep learning are sufficiently disparate fields that there is little risk of terminological confusion. Similarly, the notion of semantics introduced here is distinct from semantics in the theory of programming languages.

Although game theory was originally developed to model human interactions \cite{vonneumann:44}, it has been pointed out that it may be more directly applicable to interacting populations of algorithms, so-called \emph{machina economicus} \cite{lay:10,abernethy:11a,storkey:11,parkes:15,frongillo:15,syrgkanis:15}. This paper goes one step further to propose that games played over first-order communication protocols are a key component of the foundations of deep learning.

A source of inspiration for the essay is Bayesian networks and Markov random fields.  Probabilistic graphical models and factor graphs provide simple, powerful ways to encode a multivariate distribution's independencies into a diagram \cite{pearl:88,kschischang:01,wainwright:08}. They have greatly facilitated the design and analysis of probabilistic algorithms. However, there is no comparable framework for distributed optimization and deep learning. The essay is intended as a first step in this direction.

\section{Semantics and Representations}
\label{sec:reps}

This section defines semantics and representations. In short, the semantics of a function is how it categorizes its inputs; a function is a representation if it is selected to optimize an objective. The connection between the definition of representation below and ``representation learning'' is clarified in section~\ref{sec:ebp}.

Possible world semantics was introduced by Lewis to formalize the meaning of sentences in terms of counterfactuals \cite{lewis:86}. Let $\cP$ be a proposition about the world. Its truth depends on its content and the state of the world. Rather than allowing the state of the world to vary, it is convenient to introduce the set $W$ of all possible worlds. 

Let us denote proposition $\cP$ applied in world $w\in W$ by $\cP(w)$. The meaning of $\cP$ is then the mapping $v_\cP:W\rightarrow\{0,1\}$ which assigns 1 or 0 to each $w\in W$ according to whether or not proposition $\cP(w)$ is true. Equivalently, the meaning of the proposition is the ordered pair consisting of: all worlds, and the subset of worlds where it is true:
\begin{equation}
	\label{e:pw2}
	\underbrace{W}_{\text{set of possible worlds}} 
	\supset \underbrace{v_{\cP}^{-1}(1)}_{\text{subset of worlds where $\cP$ is true}}
\end{equation}
For example, the meaning of $\cP_{blue}(\text{\emph{that}})=$``\emph{that} is blue'' is the subset $v_{\cP_{blue}}^{-1}(1)$ of possible worlds where I am pointing at a blue object. The concept of blue is rendered explicit in an exhaustive list of  possible examples.

A simple extension of possible world semantics from propositions to arbitrary functions is as follows \cite{balduzzi:11ffp}:

\begin{defn}[semantics]\label{d:meaning}\eod
	Given function $f:X\rightarrow Y$, the \textbf{semantics} or \textbf{meaning} of output $y\in Y$ is the ordered pair of sets 
	\begin{equation}
		\label{e:proc2}
		\underbrace{X}_{\text{set of possible inputs}}\supset \underbrace{f^{-1}(y)}_{\text{subset causing $f$ to output $y$}}
	\end{equation}
	Functions implicitly categorize inputs by assigning outputs to them; the meaning of an output is the category.
\end{defn}

Whereas propositions are true or false, the output of a function is neither. However, if two functions both optimize a criterion, then one can refer to how \emph{accurately} one function \emph{represents} the other.  Before we can define representations we therefore need to take a quick detour through optimization:

\begin{defn}[optimization problem]\label{d:opt}\eod
	An \textbf{optimization problem} is a pair $(\Theta, \fobj)$ consisting in parameter-space $\Theta\subset \bR^d$ and objective $\fobj:\Theta\rightarrow \bR$ that is differentiable almost everywhere. 

	The \textbf{solution} to the global optimization problem is
	\begin{equation}
		\btheta^* = \argopt_{\btheta\in\Theta}\fobj(\btheta),
	\end{equation}
	which is either a maximum or minimum according to the nature of the objective. 
\end{defn}

The solution may not be unique; it also may not exist unless further restrictions are imposed. Such details are ignored here.

Next recall the black-box optimization framework introduced by Nemirovski and Yudin \cite{nemirovski:83,agarwal:09,raginsky:11,arjevani:15}. 

\begin{defn}[communication protocol]\label{d:protocol}\eod
	A \textbf{communication protocol} for optimizing an unknown objective $\fobj:\Theta\rightarrow \bR$ consists in a User (or Player) and an Oracle. On each round, User presents a \textbf{query} $\btheta\in\Theta$. Oracle can \textbf{respond} in one of two ways, depending on the nature of the protocol:
	\begin{itemize}
		\item \emph{Black-box (zeroth-order) protocol.}\\
		Oracle responds with the value $\fobj(\btheta)$. \\
		
		{\small
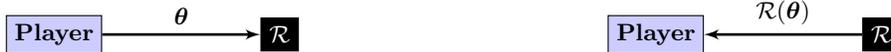
		\begin{tikzpicture}[>=latex']
		    \node[red_block] at (0,0)  (player) {\rm{\textbf{Player}}};
		    \node[yellow_block] at (3,0)  (block) {\color{white}$\fobj$};
		    \draw[->,line width=1pt] (player)   edge node[above] {$\btheta$} (block);
		    \node[red_block] at (8,0)  (player_a) {\rm{\textbf{Player}}};
		    \node[yellow_block] at (11,0)  (oracle) {\color{white}$\fobj$};
		    \draw[->,line width=1pt] (oracle) edge node[above] {$\fobj(\btheta)$} (player_a);
		\end{tikzpicture}
		}\\

		\item \emph{Gray-box (first-order) protocol.}\\
		Oracle responds with either the gradient $\grad\fobj(\btheta)$ or with the gradient together with the value.\\

		{\small
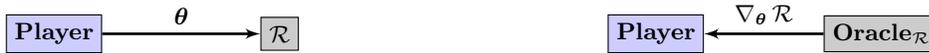
		\begin{tikzpicture}[>=latex']
		    \node[red_block] at (0,0)  (player) {\rm{\textbf{Player}}};
		    \node[block] at (3,0)  (block) {$\fobj$};
		    \draw[->,line width=1pt] (player)   edge node[above] {$\btheta$} (block);
		    \node[red_block] at (8,0)  (player_a) {\rm{\textbf{Player}}};
		    \node[o_block] at (11,0)  (oracle) {\rm{\textbf{Oracle$_{\fobj}$}}};
		    \draw[->,line width=1pt] (oracle) edge node[above] {$\grad_{\btheta}\fobj$} (player_a);
		\end{tikzpicture}
		}

	\end{itemize}
\end{defn}

The protocol specifies how Player and Oracle interact without specifying the algorithm used by Player to decide which points to query. The next section introduces \emph{distributed communication protocols} as a general framework that includes a variety of deep learning architectures as special cases -- again without specifying the precise algorithms used to perform weight updates.

Unlike \cite{nemirovski:83,raginsky:11} we do not restrict to convex problems.  Finding a global optimum is not always feasible, and in practice often unnecessary. 

\begin{defn}[representation]\label{d:foo}\eod
	Let $\cF\subset\{f:X\rightarrow Y\}$ be a function space and
	\begin{equation}
		f:\Theta  \rightarrow \cF:
		\btheta  \mapsto f_\btheta(\bullet)
	\end{equation}
	 be a map from parameter-space to functions. Further suppose that objective function $\fobj:\cF\rightarrow \bR$ is given.

	A \textbf{representation} is a local solution to the optimization problem
	\begin{equation}
		f_{\hat{\btheta}} \quad\text{where}\quad\hat{\btheta}\in \arglocopt_{\btheta\in\Theta}\fobj(f_\btheta),
	\end{equation}
	corresponding to a \emph{local} maximum or minimum according to whether the objective is minimized or maximized. 

	Intuitively, the objective quantifies the extent to which functions in $\cF$ categorize their inputs similarly. The operation $\arglocopt$ applies a first-order method to find a function whose semantics resembles the optimal solution $f_{\btheta^*}$ where $\btheta^*=\argopt_{\btheta\in\Theta}\fobj(f_{\btheta})$. 
\end{defn}

In short, representations are functions with useful semantics, where usefulness is quantifed using a specific objective: the lower the loss or higher the reward associated with a function, the more useful it is. The relation between Definition~\ref{d:foo} and representations as commonly understood in the deep learning literature is discussed in section~\ref{sec:ebp} below.

\begin{rem}[value function semantics]\label{rem:sutton}\eod
	In related work, Sutton \emph{et al} \cite{sutton:11} proposed that semantics -- i.e. knowledge about the world -- can be encoded in general value functions that provide answers to specific questions about expected rewards. Definition~\ref{d:meaning} is more general than their approach since it associates a semantics to \emph{any} function. However, the function must arise from optimizing an objective for its semantics to accurately represent a phenomenon of interest.
\end{rem}

\subsection{Supervised learning}
\label{sec:supervised}

The main example of a representation arises under supervised learning.

\begin{repn}[supervised learning]\label{rep:sup}\eod
	Let $X$ and $Y$ be an input space and a set of labels and $\loss:Y\times Y\rightarrow \bR$ be a loss function. Suppose that $\{f_\btheta:X\rightarrow Y\,|\,\theta\in\Theta\}$ is a parametrized family of functions.
	\begin{itemize}
		\item \emph{Nature}
		which samples labeled pairs $(x,y)$ i.i.d. from distribution $\bP_{XY}$, singly or in batches. 
		\item \emph{Predictor} chooses parameters
		$\btheta\in\Theta$.
		\item \emph{Objective}
			is
			\begin{equation}
				\fobj(\btheta) = \expec_{(x,y)\sim \bP_{XY}}[\loss(f_\btheta(x),y)].
			\end{equation}			
	\end{itemize}

	The query and responses phases can be depicted graphically as\\

	{\small
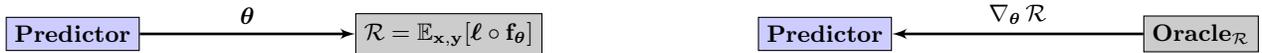
	\begin{tikzpicture}[>=latex']
	    \node[red_block] at (0,0)  (player) {\rm{\textbf{Predictor}}};
	    \node[block] at (5,0)  (block) {$\mathbf{\fobj=\expec_{x,y}[\loss\circ f_\btheta]}$};
	    \draw[->,line width=1pt] (player)   edge node[above] {$\btheta$} (block);
	    \node[red_block] at (10,0)  (player_a) {\rm{\textbf{Predictor}}};
	    \node[o_block] at (15,0)  (oracle) {\rm{\textbf{Oracle$_{\fobj}$}}};
	    \draw[->,line width=1pt] (oracle) edge node[above] {$\grad_{\btheta}\fobj$} (player_a);
	\end{tikzpicture}
	}\\

	The predictor $f_{\hat{\btheta}} = \arglocmin_{\btheta\in\Theta}\fobj(\btheta)$ is then a representation of the optimal predictor $f_{\btheta^*}=\argmin_{\btheta\in\Theta} \fobj(\btheta)$.	
\end{repn}

A commonly used mapping from parameters to functions is
\begin{equation}
	f:\Theta\rightarrow \cF:\btheta\mapsto f_\btheta(\bullet) := \langle\phi(\bullet),\btheta\rangle
\end{equation}
where a feature map $\phi:X\rightarrow \bR^d$ is fixed.

The setup admits a variety of complications in practice. Firstly, it is typically infeasible even to find a local optimum. Instead, a solution that is within some small $\epsilon>0$  of the local optimum suffices. Secondly, the distribution $\bP_{XY}$ is unknown, so the expectation is replaced by a sum over a finite sample. The quality of the resulting representation has been extensively studied in statistical learning theory \cite{vapnik:95}. Finally, it is often convenient to modify the objective, for example by incorporating a regularizer. Thus, a more detailed presentation would conclude that 
\begin{equation}
	\hat{\btheta} \approx \arglocmin_{\btheta\in\Theta} \sum_{i=1}^n \loss(f_{\btheta}(x_i),y_i) + \Omega(\btheta)
\end{equation}
yields a representation $f_{\hat{\btheta}}$ of the solution to $\argmin_{\btheta} \expec_{\bP_{XY}}[\loss(f_{\btheta}(x),y)]$. To keep the discussion and notation simple, we do not consider any of these important details.

It is instructive to unpack the protocol, by observing that the objective $\fobj$ is a composite function involving $f(\btheta,x)$, $\loss(f,y)$ and $\expec[\bullet]$: \\

	{\small
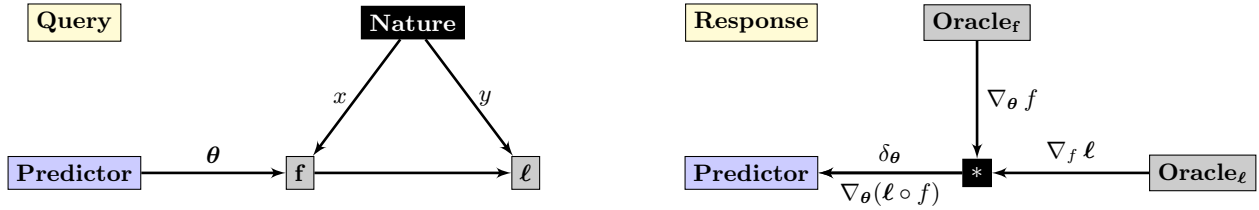
	\begin{tikzpicture}[>=latex']
		\node[wblock] at (0,0) (glabel) {\textbf{Query}};
	    \node[yellow_block] at (4.5,0) (unknown_in) { \color{white}\rm{\textbf{Nature}}};
	    \node[red_block] at (0,-2)  (player) {\rm{\textbf{Predictor}}};
	    \node[block] at (3,-2)  (block1) {$\mathbf{f}$};
	    \node[block] at (6,-2)  (block2) {$\loss$};
	    \draw[->,line width=1pt] (unknown_in)   edge node[left] {$x$} (block1);
	    \draw[->,line width=1pt] (player)   edge node[above] {$\btheta$} (block1);
	    \draw[->,line width=1pt] (block1)   edge node[left] {$\,$} (block2);
	    \draw[->,line width=1pt] (unknown_in)   edge node[right] {$y$} (block2);
	
	    \node[wblock] at (9,0) (glabel) {\textbf{Response}};
	    \node[red_block] at (9,-2)  (player_out) {\rm{\textbf{Predictor}}};
	    \node[o_block] at (15,-2)  (oracle_l)    {\rm{\textbf{Oracle$_{\loss}$}}};
	    \node[o_block] at  (12,0) (oracle_f)     {\rm{\textbf{Oracle$\mathbf{_{f}}$}}};
	    \node[green_block] at (12,-2)  (block) {$\color{white}*$};

	    \draw[->,line width=1pt] (oracle_l) edge node[above] {$\grad_{f}\loss$} (block);
	    \draw[->,line width=1pt] (oracle_f) edge node[right] {$\grad_{\btheta}f$} (block);
	    \draw[->,line width=1pt] (block) edge node[below] {$\grad_{\btheta}(\loss\circ f)$} (player_out);
	    \draw[->,line width=1pt] (block) edge node[above] {$\delta_\btheta$} (player_out);
	\end{tikzpicture}
	}

The notation $\delta_\btheta$ is borrowed from backpropagation. It is shorthand for the derivative of the objective with respect to parameters $\btheta$.

Nature is not a deterministic black-box since it is not queried directly: Nature produces $(x,y)$ pairs stochastically, rather than in response to specific inputs. Our notion of black-box can be extended to stochastic black-boxes, see e.g. \cite{schulman:15}. However, once again we prefer to keep the exposition as simple as possible.

\subsection{Unsupervised learning}
\label{sec:unsupervised}

The second example concerns fitting a probabilistic or generative model to data. A natural approach is to find the distribution under which the observed data is most likely:

\begin{repn}[maximum likelihood estimation]\eod
	Let $X$ be a data space.
	\begin{itemize}
		\item \emph{Nature}
			samples points from distribution $\bP_X$.
		\item \emph{Estimator}
			chooses parameters $\btheta\in\Theta$.
		\item \emph{Operator} $\bQ(x;\btheta)=\bQ_{\btheta}(x)$ computes a probability density on $X$ that depends on parameter $\theta$.
		\item \emph{Operator} $-\log(\cdot)$ acts as a loss. The objective is to mimimize
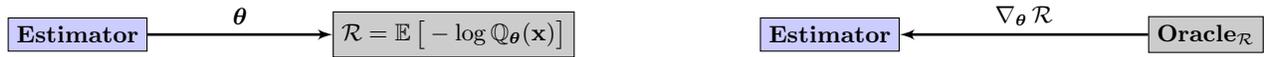
			\begin{equation}
				\fobj(\btheta) 
				 := -\expec_{x\sim\bP_X}\big[\log \bQ_\btheta(x)\big].				
			\end{equation}
	\end{itemize}

	{\small
	\begin{tikzpicture}[>=latex']
	    \node[red_block] at (0,0)  (player) {\rm{\textbf{Estimator}}};
	    \node[block] at (5,0)  (block) {$\mathbf{\fobj=\expec\big[-\log \bQ_\btheta(x)\big]}$};
	    \draw[->,line width=1pt] (player)   edge node[above] {$\btheta$} (block);
	    \node[red_block] at (10,0)  (player_a) {\rm{\textbf{Estimator}}};
	    \node[o_block] at (15,0)  (oracle) {\rm{\textbf{Oracle$_{\fobj}$}}};
	    \draw[->,line width=1pt] (oracle) edge node[above] {$\grad_{\btheta}\fobj$} (player_a);
	\end{tikzpicture}
	}\\

\end{repn}

The estimate $\bQ(x;\hat{\btheta})$, where $\hat{\btheta}\in \arglocmin_{\btheta\in\Theta}\fobj(\btheta)$, is a representation of the optimal solution, and can also be considered a representation of $\bP_X$. The setup extends easily to maximum \emph{a posteriori} estimation. 

As for supervised learning, the protocol can be unpacked by observing that the objective has a compositional structure:\\

	{\small
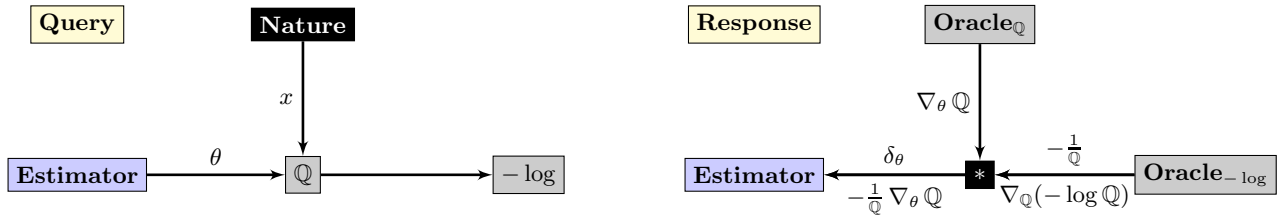
	\begin{tikzpicture}[>=latex']
		\node[wblock] at (0,0) (glabel) {\textbf{Query}};
	    \node[yellow_block] at (3,0) (unknown_in) {\color{white}\rm{\textbf{Nature}}};
	    \node[red_block] at (0,-2)  (player)      {\rm{\textbf{Estimator}}};
	    \node[block] at (3,-2)  (block1) {$\bQ$};
	    \node[block] at (6,-2)  (block2) {$\mathbf{-\log}$};
	    \draw[->,line width=1pt] (unknown_in)   edge node[left] {$x$} (block1);
	    \draw[->,line width=1pt] (player)   edge node[above] {$\theta$} (block1);
	    \draw[->,line width=1pt] (block1)   edge node[left] {$\,$} (block2);
	
	    \node[wblock] at (9,0) (glabel) {\textbf{Response}};
	    \node[red_block] at  (9,-2) (player_out) {\rm{\textbf{Estimator}}};
	    \node[o_block] at (15,-2)  (oracle_l)    {\rm{\textbf{Oracle$_{-\log}$}}};
	    \node[o_block] at (12,0)  (oracle_f)     {\rm{\textbf{Oracle$_{\bQ}$}}};
	    \node[green_block] at (12,-2)  (block) {$\color{white}*$};

	    \draw[->,line width=1pt] (oracle_l) edge node[above] {$-\frac{1}{\bQ}$} (block);
	    \draw[->,line width=1pt] (oracle_l) edge node[below] {$\grad_\bQ(-\log \bQ)$} (block);
	    \draw[->,line width=1pt] (oracle_f) edge node[left] {$\grad_{\theta}\bQ$} (block);
	    \draw[->,line width=1pt] (block) edge node[below] {$-\frac{1}{\bQ}\grad_\theta\bQ$} (player_out);
	    \draw[->,line width=1pt] (block) edge node[above] {$\delta_\theta$} (player_out);
	\end{tikzpicture}
	}

\subsection{Reinforcement learning}
\label{sec:valprox}

The third example is taken from reinforcement learning \cite{sutton:98}. We will return to reinforcement learning in section~\ref{sec:pg}, so the example is presented in some detail. In reinforcement learning, an agent interacts with its environment, which is often modeled as a Markov decision process consisting of state space $\cS\subset\bR^m$, action space $\cA\subset\bR^d$, initial distribution $\bP_1(\sta)$ on states, stationary transition distribution $\bP(\sta_{t+1}|\sta_t,\act_t)$ and reward function $r:\cS\times \cA\rightarrow \bR$. The agent chooses actions based on a \emph{policy}: a function $\bmu_\theta:\cS\rightarrow \cA$ from states to actions. The goal is to find the optimal policy. 

Actor-critic methods break up the problem into two pieces \cite{barto:83}. The critic estimates the expected value of state-action pairs given the current policy, and the actor attempts to find the optimal policy using the estimates provided by the critic. The critic is typically trained via temporal difference methods \cite{sutton:88,dann:14}.

Let $\bP_t(\sta\rightarrow \sta',\bmu)$ denote the distribution on states $\sta'$ at time $t$ given policy $\bmu$ and initial state $\sta$ at $t=0$ and let $\rho^{\bmu}(\sta') = \int_\cS\sum_{t=0}^\infty\gamma^t\bP_1(\sta)\bP_t(\sta\rightarrow \sta',\bmu)d\sta$. Let $r_t^\gamma = \sum_{\tau=t}^\infty\gamma^{\tau-t} r(\sta_\tau,\act_\tau)$ be the discounted future reward. Define the value of a state-action pair as
\begin{equation}
	 Q^{\bmu}(\sta,\act) = \expec[r_1^\gamma | {\mathbf S}_1=\sta,{\mathbf A}_1=\act;\bmu].
\end{equation}
Unfortunately, the value-function $Q^{\bmu}(\sta,\act)$ cannot be queried. Instead, temporal difference methods take a bootstrapped approach by minimizing the Bellman error:
\begin{equation}
	\label{e:MSBE}
	\loss_{BE}(\vt) = 
	\expec_{(\sta,\act)\sim (\rho^{\bmu}, \bmu)}\Big[\Big(
	r(\sta,\act) + \gamma Q^{\vt}(\sta',\bmu(\sta')) - Q^\vt(\sta,\act)
	\Big)^2\Big]
\end{equation}
where $\sta'$ is the state subsequent to $\sta$.

\begin{repn}[temporal difference learning]\eod
	Critic interacts with black-boxes Actor and Nature.\footnote{Nature's outputs depend on Actor's actions, so the Query graph should technically have an additional arrow from Actor to Nature. 
	}
	\begin{itemize}
		\item \emph{Critic} plays parameters $\vt$.
		\item \emph{Operator} $Q$ and $\loss_{BE}$ estimates the value function and compute the Bellman error. In practice, it turns out to \emph{clone} the value-estimate periodically and compute a slightly modified Bellman error:
		\begin{equation}
			\loss_{BE}(\vt) = 
			\expec_{(\sta,\act)\sim (\rho^{\bmu}, \bmu)}\Big[\Big(
			r(\sta,\act) + \gamma Q^{\tilde{\vt}}(\sta',\bmu(\sta')) - Q^\vt(\sta,\act)
			\Big)^2\Big]
		\end{equation}
		where $Q^{\tilde{\vt}}$ is the cloned estimate. Cloning improves the stability of TD-learning \cite{Mnih:2015wq}. A nice conceptual side-effect of cloning is that TD-learning reduces to gradient descent.\\

		{\small
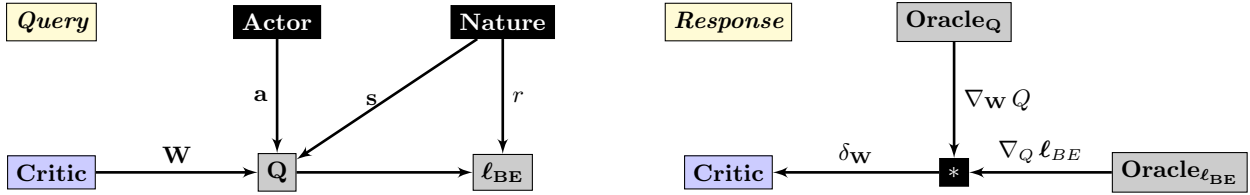
		\begin{tikzpicture}[>=latex']
	    	\node[wblock] at (0,0) (glabel) {\textbf{Query}};
	    	\node[yellow_block] at (6,0) (unknown_in)  {\color{white}\rm{\textbf{Nature}}};
	    	\node[yellow_block] at (3,0) (actor)       {\color{white}\rm{\textbf{Actor}}};
	    	\node[red_block] at (0,-2)  (player) {\rm{\textbf{Critic}}};
	    	\node[block] at (3,-2)  (block1) {$\mathbf{Q}$};
	    	\node[block] at (6,-2)  (block2) {$\mathbf{\loss_{BE}}$};
	    	\draw[->,line width=1pt] (unknown_in)   edge node[left] {$\sta$} (block1);
	    	\draw[->,line width=1pt] (player)   edge node[above] {$\Wt$} (block1);
	    	\draw[->,line width=1pt] (block1)   edge node[left] {$\,$} (block2);
	    	\draw[->,line width=1pt] (unknown_in)   edge node[right] {$r$} (block2);
	    	\draw[->,line width=1pt] (actor) edge node[left] {$\act$} (block1);
		
	    	\node[wblock] at (9,0) (glabel) {\textbf{Response}};
	    	\node[red_block] at (9,-2)  (player_out) {\rm{\textbf{Critic}}};
	    	\node[o_block] at (15,-2)  (oracle_l) {\rm{\textbf{Oracle$\mathbf{_{\loss_{BE}}}$}}};
	    	\node[o_block] at  (12,0) (oracle_f)  {\rm{\textbf{Oracle$\mathbf{_{Q}}$}}};
	    	\node[green_block] at (12,-2)  (block) {$\color{white}*$};
	
	    	\draw[->,line width=1pt] (oracle_l) edge node[above] {$\grad_{Q}\loss_{BE}$} (block);   	
	    	\draw[->,line width=1pt] (oracle_f) edge node[right] {$\grad_{\Wt}Q$} (block);
	    	\draw[->,line width=1pt] (block) edge node[above] {$\delta_\Wt$} (player_out);
		\end{tikzpicture}
		}\\
	\end{itemize}
	The estimate is a representation of the true value function.
\end{repn}

\begin{rem}[on temporal difference learning as first-order method]\eod
	Temporal difference learning is not strictly speaking a gradient-based method \cite{dann:14}. The residual gradient method performs gradient descent on the Bellman error, but suffers from double sampling \cite{baird:95}. Projected fixpoint methods minimize the \emph{projected} Bellman error via gradient descent and have nice convergence properties \cite{sutton:09,sutton:09a,maei:10}. An interesting recent proposal is implicit TD learning \cite{tamar:14},  which is based on implicit gradient descent \cite{toulis:14}.
\end{rem}

Section~\ref{sec:pg} presents the Deviator-Actor-Critic model which simultaneously learns a value-function estimate and a locally optimal policy.

\section{Protocols and Grammars}
\label{sec:comp}

It is often useful to decompose complex problems into simpler subtasks that can handled by specialized modules. Examples include variational autoencoders, generative adversarial networks and actor-critic models. Neural networks are particularly well-adapted to modular designs, since units, layers and even entire networks can easily be combined analogously to bricks of lego \cite{bottou:91}. 

However, not all configurations are viable models. A methodology is required to distinguish good designs from bad. This section provides a basic language to describe how bricks are glued together that may be a useful design tool. The idea is to extend the definitions of optimization problems, protocols and representations from section~\ref{sec:reps} from single to multi-player optimization problems. 

\begin{defn}[game]\label{d:game}\eod
	A \textbf{distributed optimization problem} or \textbf{game} $\big([N],\Theta, \loss)$ is a set $[N]=\{1,\ldots N\}$ of players, a parameter space $\Theta = \prod_{i=1}^N\Theta_i$, and loss vector $\loss=(\ell_1,\ldots, \ell_N):\Theta\rightarrow \bR^N$. Player $i$ picks moves from $\Theta_i\subset \bR^{d_i}$ and incurs loss determined by $\ell_i:\Theta\rightarrow \bR$. The goal of each player is to minimize its loss, which depends on the moves of the other players. 
\end{defn}

The classic example is a \emph{finite game} \cite{vonneumann:44}, where player $i$ has a menu of $d_i$-actions and chooses a distribution over actions, $\btheta_i\in\Theta_i=\triangle_{d_i}=\{ (\theta_1,\ldots, \theta_{d_i}):\sum_{j=1}^{d_i} \theta_j=1\text{ and } \theta_j\geq 0\}$ on each round. Losses are specified for individual actions, and extended linearly to distributions over actions. A natural generalization of finite games is \emph{convex games} where the parameter spaces are compact convex sets and each loss $\ell_i$ is a convex function in its $i^{\text{th}}$-argument \cite{stoltz:07}. It has been shown that players implementing no-regret algorithms are guaranteed to converge to a correlated equilibrium in convex games \cite{foster:97,blum:07,stoltz:07}.

The notion of game in Definition~\ref{d:game} is too general for our purposes. Additional structure is required.
\begin{defn}[computation graph]\label{def:cg}\eod
	A \textbf{computation graph} is a directed acyclic graph with two kinds of nodes:
	\begin{itemize}
		\item \emph{Inputs} 
		are set externally (in practice by Players or Oracles).
		\item \emph{Operators} 
		produce outputs that are a fixed function of their parents' outputs.
	\end{itemize}	
\end{defn}

Computation graphs are a useful tool for calculating derivatives \cite{griewank:08,bergstra:10,bastien:12,merrienboer:15,baydin:14}. For simplicity, we restrict to deterministic computation graphs. More general stochastic computation graphs are studied in \cite{schulman:15}.

A \emph{distributed} communication protocol extends the communication protocol in Definition~\ref{d:protocol} to multiplayer games using two computation graphs. 

\begin{defn}[distributed communication protocol]\label{d:dcp}\eod
	A \textbf{distributed communication protocol} is a game where each round has two phases, determined by two computation graphs:
	\begin{itemize}
		\item \emph{Query phase.} 
		Players provide inputs to the Query graph \emph{($\quer$)} that Operators transform into outputs.
		\item \emph{Response phase.} 
		Operators in $\quer$ act as Oracles in the Response graph \emph{($\resp$)}: they input subgradients that are transformed and communicated to the Players.
	\end{itemize}
	The moves chosen by Players depend only on their prior moves and the information communicated to them by the Response graph.
\end{defn}

The protocol specifies how Players and Oracles communicate without specifying the optimization algorithms used by the Players.  The addition of a Response graph allows more general computations than simply backpropagating the gradients of the Query phase. The additional flexibility allows the design of new algorithms, see sections~\ref{sec:pg} and \ref{sec:kb} below. It is also sometimes necessary for computational reasons. For example, backpropagation through time on recurrent networks typically runs over a truncated Response graph \cite{elman:90,williams:90,williams:95}.

Suppose that we wish to optimize an objective function $\fobj:\Theta\rightarrow\bR$ that depends on all the moves of all the players. Finding a global optimum is clearly not feasible. However, we may be able to construct a protocol such that the players are jointly able to find local optima of the objective. In such cases, we refer to the protocol as a grammar:

\begin{defn}[grammar]\label{d:grammar}\eod
	A \textbf{grammar} for objective $\fobj:\Theta\rightarrow\bR$ is a distributed communication protocol where the Response graph provides \emph{sufficient} first-order information to find a local optimum of $(\fobj,\Theta)$. 
\end{defn}

The guarantee ensures that the representations constructed by Players in a grammar can be combined into a coherent distributed representation. That is, it ensures that the representations constructed by the Players transform data in a way that is useful for optimizing the shared objective $\fobj$.

The Players' losses need not be explicitly computed. All that is necessary is that the Response phase communicate the gradient information needed for Players to locally minimize their losses -- and that doing so yields a local optimum of the objective.

\vspace{2mm}
\noindent
\textbf{\sffamily{Basic building blocks: function composition ($\quer$) and the chain rule ($\resp$).}}
Functions can be inserted into grammars as lego-like building blocks via function composition during queries and the chain rule during responses. Let $G(\btheta,F)$ be a function that takes inputs $\btheta$ and $F$, provided by a Player and  by upstream computations respectively. The output of $G$ is communicated downstream in the Query phase:\\
	
	{\small
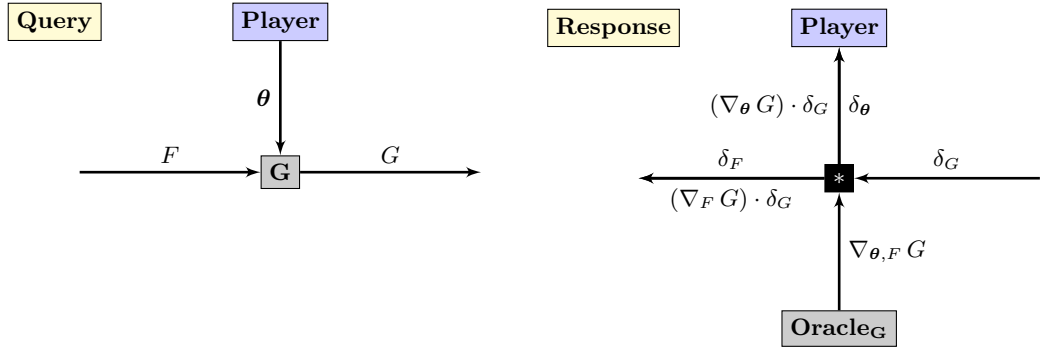
	\begin{tikzpicture}[>=latex']
		\node[wblock] at (0,0) (glabel) {\textbf{Query}};
	    \node[no_block] at (0,-2) (unknown_in) {$\,$};
	    \node[no_block] at (6,-2) (unknown_out) {$\,$};
	    \node[red_block] at (3,0)  (player) {\rm{\textbf{Player}}};
	    \node[no_block] at (0,-4) (unknown_x) {$\,$};
	    \node[block] at (3,-2)  (block) {$\mathbf{G}$};
	    \draw[->,line width=1pt] (unknown_in)   edge node[above] {$F$} (block);
	    \draw[->,line width=1pt] (player)   edge node[left] {$\btheta$} (block);
	    \draw[->,line width=1pt] (block)   edge node[above] {$G$} (unknown_out);
	\end{tikzpicture}
	}	
	{\small
	\begin{tikzpicture}[>=latex']
		\node[wblock] at (0,0) (glabel) {\textbf{Response}};
	    \node[no_block] at (0,-2) (unknown_in) {$\,$};
	    \node[no_block] at (6,-2) (unknown_out) {$\,$};
	    \node[red_block] at (3,0)  (player) {\rm{\textbf{Player}}};
	    \node[green_block] at (3,-2)  (block) {$\color{white}*$};
	    \node[o_block] at (3,-4)  (oracle) {\rm{\textbf{Oracle$\mathbf{_G}$}}};
	    \draw[->,line width=1pt] (block)   edge node[left] {$(\grad_\btheta G)\cdot \delta_{G}$} (player);
	    \draw[->,line width=1pt] (block)   edge node[right] {$\delta_{\btheta}$} (player);
	    \draw[->,line width=1pt] (block)   edge node[above] {$\delta_{F}$} (unknown_in);
	    \draw[->,line width=1pt] (block)   edge node[below] {$(\grad_F G)\cdot \delta_{G}$} (unknown_in);
	    \draw[->,line width=1pt] (unknown_out)   edge node[above] {$\delta_{G}$} (block);
	    \draw[->,line width=1pt] (oracle)   edge node[right] {$\grad_{\btheta,F}G$} (block);
	\end{tikzpicture}
	}\\

	The chain rule is implemented in the Response phase as follows. Oracle$_G$ reports the gradient $\grad_{\btheta,F}G:= (\grad_\btheta G,\grad_F G)$ in the Response phase. Operator ``$*$'' computes the products $(\grad_\btheta G\cdot \delta_G,\grad_F G\cdot \delta_G)$ via matrix multiplication. The projection of the product onto the first and second components\footnote{Alternatively, to avoid having ``$*$'' produce two outputs, the entire vector can be reported in both direction with the irrelevant components ignored.} are reported to Player and upstream respectively.

\vspace{2mm}
\noindent
\textbf{\sffamily{Summary of guarantees.}}
A selection of examples are presented below. Guarantees fall under the following broad categories:
\begin{enumerate}
	\item \emph{Exact gradients.}\\
	Under error backpropagation the Response graph implements the chain rule, which guarantees that Players receive the gradients of their loss functions; see section~\ref{sec:ebp}.
	\item \emph{Surrogate objectives.} \\
	The variational autoencoder uses a surrogate objective: the variational lower bound. Maximizing the surrogate is guaranteed to also maximize the true objective, which is computational intractable; see section~\ref{sec:vae}.
	\item \emph{Learned objectives.} \\
	In the case of generative adversarial network and the DAC-model, some of the players learn a loss that is guaranteed to align with the true objective, which is unknown; see sections~\ref{sec:advers} and \ref{sec:pg}.
	\item \emph{Estimated gradient.}\\
	In the DAC-model and kickback, gradient estimates are substituted for the true gradient; see sections~\ref{sec:pg} and \ref{sec:kb}. Guarantees are provided on the estimates.
\end{enumerate}

\begin{rem}[fine- and coarse-graining]\eod
	There is considerable freedom regarding the choice of players. In the examples below, players are typically chosen to be layers or entire neural networks to keep the diagrams simple. It is worth noting that zooming in, such that players correspond to individual units, has proven to be a useful tool when analyzing neural networks \cite{bvb:15,doco:15,gradprop:15}.  

	The game-theoretic formulation is thus scale-free and can be coarse- or fine-grained as required. A mathematical language for tracking the structure of hierarchical systems at different scales is provided by operads, see \cite{spivak:13} and the references therein, which are the natural setting to study the composition of operators that receive multiple inputs.
\end{rem}

\subsection{Error backpropagation}
\label{sec:ebp}

The main example of a grammar is a neural network using error backpropagation to perform supervised learning. Layers in the network can be modeled as players in a game. Setting each (p)layer's objective as the network's loss, which it minimizes using gradient ascent, yields backpropagation.

\begin{coupling}[backpropagation]\label{c:chain_rule}\eod
	An $L$-layer neural network can be reformulated as a game played between $L+1$ players, corresponding to \emph{Nature} and the \emph{Layers} of the network. The query graph for a 3-layer network is:\\

	\vspace{3mm}
	{\small
	\begin{tikzpicture}[>=latex']
		\node[wblock] at (0,0) (glabel) {\textbf{Query}};	    
    	\node[yellow_block] at (0,-2) (player_env1) {\color{white}\rm{\textbf{Nature}}};
    	\node[red_block] at (3,0)  (player1) {\rm{\textbf{Layer$_1$}}};
    	\node[red_block] at (7,0)  (player2) {\rm{\textbf{Layer$_2$}}};
    	\node[red_block] at (11,0) (player3) {\rm{\textbf{Layer$_3$}}};
    	\node[yellow_block] at (15,0) (player_env2) {\color{white}\rm{\textbf{Nature}}};
    	\node[block] at (3,-2)  (block1) {$\mathbf{S_1}$};
    	\node[block] at (7,-2)  (block2) {$\mathbf{S_2}$};
    	\node[block] at (11,-2) (block3) {$\mathbf{S_3}$};
    	\node[block] at (15,-2) (block4) {$\loss$};
    	\draw[->,line width=1pt] (player_env1) --                        (block1);
    	\draw[->,line width=1pt] (player1)   edge node[left] {$\btheta_1$} (block1);
    	\draw[->,line width=1pt] (player2)   edge node[left] {$\btheta_2$} (block2);
    	\draw[->,line width=1pt] (player3)   edge node[left] {$\btheta_3$} (block3);
    	\draw[->,line width=1pt] (player_env2)   edge node[left] {$y$}     (block4);
    	\draw[->,line width=1pt] (block1) -- (block2);
    	\path[->,line width=1pt] (player_env1) edge node[above] {$x$} (block1);
    	\path[->,line width=1pt] (block1) edge node[above] {$S_1$} (block2);
    	\path[->,line width=1pt] (block2) edge node[above] {$S_2$} (block3);    
    	\path[->,line width=1pt] (block3) edge node[above] {$S_3$} (block4);    
	\end{tikzpicture}
	}
	\begin{itemize}
		\item \emph{Nature} plays samples datapoints $(x,y)$ i.i.d. from $\bP_{X\times Y}$ and acts as the zeroth player.
		\item \emph{Layer$_i$} plays weight matrices $\btheta_i$.
		\item \emph{Operators} compute $S_i(\btheta_i,S_{i-1}) := S_i(\btheta_i\cdot S_{i-1})$ for each layer, along with loss $\loss(S_L,y)$.		
	\end{itemize}

	\noindent
	The response graph performs error backpropagation:\\

	\vspace{3mm}
	{\small
	\begin{tikzpicture}[>=latex']
    	\node[wblock] at (0,0) (glabel) {\textbf{Response}};	    
    	\node[no_block] at (0,-2) (oracle_env1) {$\,$};
    	\node[o_block] at (3,-4)  (oracle1) {\rm{\textbf{Oracle$\mathbf{_1}$}}};
    	\node[o_block] at (7,-4)  (oracle2) {\rm{\textbf{Oracle$\mathbf{_2}$}}};
    	\node[o_block] at (11,-4) (oracle3) {\rm{\textbf{Oracle$\mathbf{_3}$}}};
    	\node[o_block] at (15,-2) (oracle_env2) {\rm{\textbf{Oracle$_{\loss}$}}};
    	\node[green_block] at (3,-2)  (block1) {$\color{white}\mathbf{*}$};
    	\node[green_block] at (7,-2)  (block2) {$\color{white}\mathbf{*}$};
    	\node[green_block] at (11,-2) (block3) {$\color{white}\mathbf{*}$};
    	
    	\node[red_block] at (3,0)  (player1) {\rm{\textbf{Layer$\mathbf{_1}$}}};
    	\node[red_block] at (7,0)  (player2) {\rm{\textbf{Layer$\mathbf{_2}$}}};
    	\node[red_block] at (11,0) (player3) {\rm{\textbf{Layer$\mathbf{_3}$}}};
    	\draw[->,line width=1pt] (oracle1)   edge node[left] {$\grad_{\btheta_1}S_1$} (block1);
    	\draw[->,line width=1pt] (oracle2)   edge node[left] {$\grad_{\btheta_2,S_1}S_2$} (block2);
    	\draw[->,line width=1pt] (oracle3)   edge node[left] {$\grad_{\btheta_3,S_2}S_3$}  (block3);
    	\draw[->,line width=1pt] (oracle_env2)   edge node[below] {$\grad_{S_3}\loss$} (block3);
    	\draw[->,line width=1pt] (oracle_env2)   edge node[above] {$\delta_{S_3}$} (block3);
    	\draw[->,line width=1pt] (block2) -- (block1);
    	\path[->,line width=1pt] (block2) edge node[above] {$\delta_{S_1}$} (block1);
    	\path[->,line width=1pt] (block2) edge node[below] {$(\grad_{S_1}S_2)\cdot \delta_2$} (block1);  
    	\path[->,line width=1pt] (block3) edge node[above] {$\delta_{S_2}$} (block2);    
    	\path[->,line width=1pt] (block3) edge node[below] {$(\grad_{S_2}S_3)\cdot \delta_3$} (block2);  
    	\draw[->,line width=1pt] (block1) edge node[left] {$(\grad_{\btheta_1}S_1)\cdot \delta_{S_1}$} (player1);
    	\draw[->,line width=1pt] (block1) edge node[right] {$\delta_{\btheta_1}$} (player1);
    	\draw[->,line width=1pt] (block2) edge node[left] {$(\grad_{\btheta_2}S_2)\cdot \delta_{S_2}$} (player2);
    	\draw[->,line width=1pt] (block2) edge node[right] {$\delta_{\btheta_2}$} (player2);
    	\draw[->,line width=1pt] (block3) edge node[left] {$(\grad_{\btheta_3}S_3)\cdot \delta_{S_3}$} (player3);
    	\draw[->,line width=1pt] (block3) edge node[right] {$\delta_{\btheta_3}$} (player3);
	\end{tikzpicture}
	}\\
\end{coupling}

\noindent 
The protocol can be extended to convolutional networks by replacing the matrix multiplications performed by each operator, $S_i(\btheta_i\cdot S_{i-1})$, with convolutions and adding parameterless max-pooling operators \cite{lecun:98}.

\begin{guarantee}
	The loss of every (p)layer is
	\begin{equation}
		\ell(\btheta,x,y) = \loss_y\circ S_{\btheta_L}\circ \cdots \circ S_{\btheta_1}(x)
		\quad\text{where}\quad \loss_y(\bullet):= \loss(\bullet, y)
		\quad\text{where}\quad S_{\btheta_i}(\bullet):= S_i(\btheta_i\cdot \bullet).
	\end{equation}
	It follows by the chain rule that $\resp$ communicates $\grad_{\btheta_i}\ell$ to player $i$.
\end{guarantee}

\noindent
\textbf{\sffamily{Representation learning.}}
We are now in a position to relate the notion of representation in definition~\ref{d:foo} with the standard notion of representation learning in neural networks. In the terminology of section~\ref{sec:reps}, each player learns a representation. The representations learned by the different players form a coherent distributed representation because they jointly optimize a single objective function. 

Abstractly, the objective can be written as
\begin{equation}
	\fobj(\btheta_1,\ldots, \btheta_L) = \expec_{(x,y)\sim \bP_{XY}}\Big[\loss\big(S(\btheta_1,\ldots,\btheta_L,x),y\big)\Big],	
\end{equation}
where $S(\btheta_1,\ldots,\btheta_L,x) = S_{\btheta_L}\circ \cdots \circ S_{\btheta_1} (x)$. The goal is to minimize the composite objective.

If we set $\hat{\btheta}_{1:L} \in \arglocmin_{(\btheta_1,\ldots,\btheta_L)\in\Theta} \fobj(\btheta_1,\ldots,\btheta_L)$ then the function $S_{\hat{\btheta}_{1:L}}:X\rightarrow Y$ fits the definition of representation above. Moreover, the compositional structure of the network implies that $S_{\hat{\btheta}_{1:L}}$ is composed of subrepresentations corresponding to the optimizations performed by the different players in the grammar: each function $S_{\hat{\btheta}_j}(\bullet)$ is a local optimum -- where $\hat{\btheta}_j\in \arglocmin_{\btheta_j\in\Theta_j} \fobj(\hat{\btheta}_1,\ldots,\btheta_j,\ldots, \hat{\btheta}_L)$ is optimized to transform its inputs into a form that is useful to network as a whole.

\vspace{2mm}
\noindent
\textbf{\sffamily{Detailed analysis of convergence rates.}}
Little can be said in general about the rate of converge of the layers in a neural network since the loss is not convex. However, neural networks can be decomposed further by treating the individual units as players. When the units are linear or rectilinear, it turns out that the network is a \emph{circadian game}. The circadian structure provides a way to convert results about the convergence of convex optimization methods into results about the global convergence a rectifier network to a local optimum, see \cite{doco:15}.

\subsection{Variational autoencoders}
\label{sec:vae}

The next example extends the unsupervised setting described in section~\ref{sec:unsupervised}. Suppose that observations $\{\x^{(i)}\}_{i=1}^N$ are sampled i.i.d. from a two-step stochastic process: a latent value $\z^{(i)}$ is sampled from $\bP(\z)$, after which $\x^{(i)}$ is sampled from $\bP(\x|\z^{(i)})$. 

The goal is to (i) find the maximum likelihood estimator for the observed data and (ii) estimate the posterior distribution on $\z$ conditioned on an observation $\x$. A straightforward approach is to maximize the marginal likelihood
\begin{equation}
	\label{eq:evidence}
	\btheta^* := \argmax_\btheta \prod_{i=1}^N\bQ_\btheta(\x^{(i)}), 
	\text{ where }\bQ_\btheta(\x)  = \int \bQ_\btheta(\x|\z)\bQ_\btheta(\z)d\z,
\end{equation}
and then compute the posterior
\begin{equation}
	\bQ_{\btheta^*}(\z|\x) = \frac{\bQ_{\btheta^*}(\x|\z)\bQ_{\btheta^*}(\z)}{\bQ_{\btheta^*}(\x)}.
\end{equation}
However, the integral in Eq.~\eqref{eq:evidence} is typically untractable, so a more roundabout tactic is required. The approach proposed in \cite{kingma:14} is to construct two neural networks, a decoder $\decoder_\btheta(\x|\z)$ that learns a generative model approximating $\bP(\x|\z)$, and an encoder $\encoder_\bphi(\z|\x)$ that learns a recognition model or posterior approximating $\bP(\z|\x)$. 

It turns out to be useful to replace the encoder with a deterministic function, $G_\bphi(\bepsilon,\x)$, and a noise source, $\bP_{noise}(\bepsilon)$ that are compatible. Here, compatible means that sampling $\tilde{\z}\sim \encoder_\bphi(\z|\x)$ is equivalent to sampling $\bepsilon\sim \bP_{noise}(\bepsilon)$ and computing $\tilde{\z}:= G_\bphi(\bepsilon,\x)$.

\begin{coupling}[variational autoencoder]\label{c:vlb}\eod
	A variational autoencoder is a game played between Encoder, Decoder, Noise and Environment. The query graph is\\

	\vspace{3mm}		
	{\small
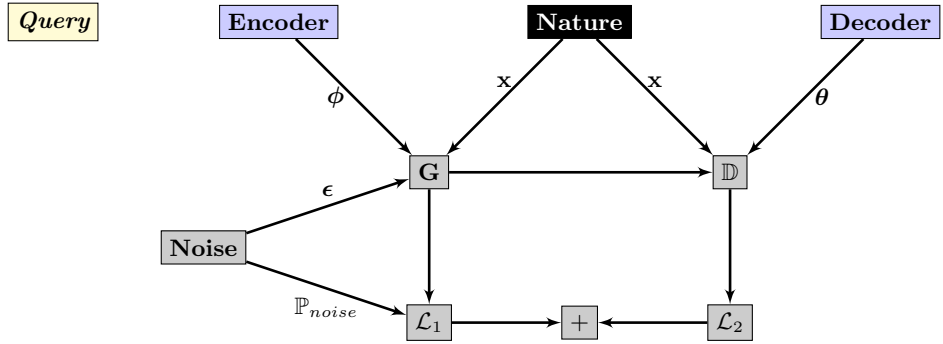
	\begin{tikzpicture}[>=latex']
		\node[wblock] at (0,0) (glabel) {\textbf{Query}};	    
    	\node[yellow_block] at (7,0) (player_env)   {\color{white}\rm{\textbf{Nature}}};
	    \node[block] at (2,-3) (player_noise) {\rm{\textbf{Noise}}};
	    \node[red_block] at (3,0)  (player1) {\rm{\textbf{Encoder}}};
	    \node[red_block] at (11,0)  (player2) {\rm{\textbf{Decoder}}};
	    \node[block] at (5,-2)  (block1) {$\mathbf{G}$};
	    \node[block] at (9,-2)  (block2) {$\decoder$};
	    \node[block] at (9,-4) (block3) {$\vlb_2$};
	    \node[block] at (5,-4) (block4) {$\vlb_1$};
	    \node[block] at (7,-4) (block5) {$\mathbf{+}$};
	    \draw[->,line width=1pt] (player1)   edge node[left] {$\bphi$} (block1);
	    \draw[->,line width=1pt] (player2)   edge node[right] {$\btheta$} (block2);
	    \draw[->,line width=1pt] (block1) -- (block2);
	    \draw[->,line width=1pt] (block1) -- (block4);
	    \draw[->,line width=1pt] (block3) -- (block5);
	    \draw[->,line width=1pt] (block4) -- (block5);
	    \path[->,line width=1pt] (player_env) edge node[above] {$\x$} (block1);
	    \path[->,line width=1pt] (player_env) edge node[above] {$\x$} (block2);
	    \path[->,line width=1pt] (player_noise) edge node[above] {$\bepsilon$} (block1);
	    \path[->,line width=1pt] (player_noise) edge node[below] {$\bP_{noise}$} (block4);
	    \path[->,line width=1pt] (block1) -- (block2);
	    \path[->,line width=1pt] (block2) edge node[above] {$\,$} (block3);    
	\end{tikzpicture}
	}
		
	\begin{itemize}
		\item \emph{Environment} plays i.i.d. samples from $\bP(\x)$
		\item \emph{Noise} plays i.i.d. samples from $\bP_{noise}(\bepsilon)$. It also communicates its density function $\bP_{noise}(\bepsilon)$, which is analogous to a gradient -- and the reason that \emph{Noise} is gray rather than black-box.
		\item \emph{Encoder} and \emph{Decoder}
		play parameters $\bphi$ and $\btheta$ respectively.
		\item \emph{Operator} $\z=G_\bphi(\bepsilon,\x)$ is a neural network that encodes samples into latent variables.
		\item \emph{Operator} $\decoder_\btheta(\z,\x)$ is a neural network that estimates the probability of $\x$ conditioned on $\z$.
		\item The remaining operators compute the (negative) variational lower bound
		\begin{equation}
			\label{eq:vlb2}
			\vlb(\btheta,\bphi;\x) 
			= \underbrace{\int \bP_{noise}(\bepsilon)
			\log \frac{\bP_{noise}(\bepsilon)}{\bP_{prior}(G_\bphi(\bepsilon,\x))}}_{\vlb_1}
			+ \underbrace{\expec_{\bepsilon\sim \bP_{noise}(\bepsilon)}\Big[-\log \decoder_\btheta\big(G_\bphi(\bepsilon,\x),\x\big)\Big]}_{\vlb_2}.
		\end{equation}
	\end{itemize}		

	\noindent
	The response graph implements backpropagation: \\

	{\small
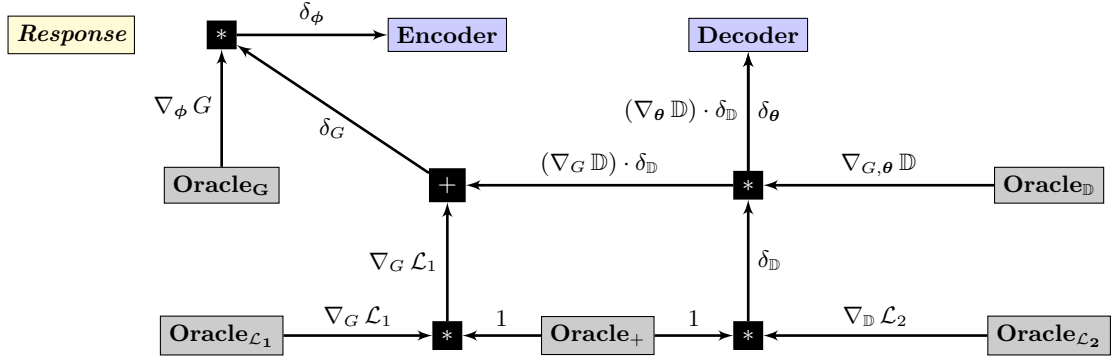
	\begin{tikzpicture}[>=latex']
	    \node[wblock] at (0,0) (glabel) {\textbf{Response}};	    
    	\node[o_block] at (2,-2)  (oracle1) {\rm{\textbf{Oracle$\mathbf{_G}$}}};
	    \node[o_block] at (13,-2)  (oracle2){\rm{\textbf{Oracle$\mathbf{_\decoder}$}}};
	    \node[o_block] at (2,-4) (oracle4)  {\rm{\textbf{Oracle$\mathbf{_{\vlb_1}}$}}};
	    \node[o_block] at (13,-4) (oracle3) {\rm{\textbf{Oracle$\mathbf{_{\vlb_2}}$}}};
	    \node[o_block] at (7,-4) (oracle5)  {\rm{\textbf{Oracle$\mathbf{_{+}}$}}};
	    \node[green_block] at (5,-2)  (block1) {$\color{white}\mathbf{+}$};
	    \node[green_block] at (9,-2)  (block2) {$\color{white}\mathbf{*}$};    
	    \node[green_block] at (5,-4)  (block3) {$\color{white}\mathbf{*}$};    
	    \node[green_block] at (9,-4)  (block4) {$\color{white}\mathbf{*}$};    
	    \node[green_block] at (2,0)  (block5)  {$\color{white}\mathbf{*}$};    
	    \node[red_block] at (5,0)  (player1) {\rm{\textbf{Encoder}}};
	    \node[red_block] at (9,0)  (player2) {\rm{\textbf{Decoder}}};
	
	    \draw[->,line width=1pt] (oracle1)   edge node[left] {$\grad_\bphi G$} (block5);
	    \draw[->,line width=1pt] (oracle2)   edge node[above] {$\grad_{G,\btheta} \decoder$} (block2);
	    \draw[->,line width=1pt] (block4)   edge node[right] {$\delta_{\decoder}$} (block2);
	    \draw[->,line width=1pt] (oracle3)   edge node[above] {$\grad_\decoder\vlb_2$} (block4);
	    \draw[->,line width=1pt] (oracle4)   edge node[above] {$\grad_G\vlb_1$} (block3);
	    \draw[->,line width=1pt] (oracle5)   edge node[above] {$1$} (block3);
	    \draw[->,line width=1pt] (oracle5)   edge node[above] {$1$} (block4);
	    \path[->,line width=1pt] (block1) edge node[below] {$\delta_{G}$} (block5);
	    \path[->,line width=1pt] (block2) edge node[above] {$(\grad_G\decoder)\cdot \delta_{\decoder}$} (		block1);
	    \path[->,line width=1pt] (block3) edge node[left] {$\grad_G\vlb_1$} (block1);
	    \draw[->,line width=1pt] (block5) edge node[above] {$\delta_\bphi$} (player1);
	    \draw[->,line width=1pt] (block2) edge node[left] {$(\grad_{\btheta}\decoder)\cdot \delta_{\decoder}$} (		player2);
	    \draw[->,line width=1pt] (block2) edge node[right] {$\delta_\btheta$} (player2);
	\end{tikzpicture}
	}
\end{coupling}

\begin{guarantee}
	The guarantee has two components:
	\begin{enumerate}
		\item Maximizing the variational lower bound yields (i) a maximum likelihood estimator and (ii) an estimate of the posterior on the latent variable \cite{kingma:14}.
		\item The chain rule ensures that the correct gradients are communicated to Encoder and Decoder.
	\end{enumerate}
	The first guarantee is that the surrogate objective computed by the query graph yields good solutions. The second guarantee is that the response graph communicates the correct gradients.
\end{guarantee}

\subsection{Generative-Adversarial networks}
\label{sec:advers}

A recent approach to designing generative models is to construct an adversarial game between Forger and Curator \cite{goodfellow:14}. Forger generates samples; Curator aims to discriminate the samples produced by Forger from those produced by Nature. Forger aims to create samples realistic enough to fool Curator.

If Forger plays parameters $\btheta$ and Curator plays $\bphi$ then the game is described succinctly via
\begin{equation}
	\arglocmin_{\btheta}\;\arglocmax_{\bphi} 
	\left[\expec_{\x\sim \bP(\x)}\big[\log D_\bphi(\x)\big]
	+ \expec_{\bepsilon\sim\bP_{noise}(\bepsilon)}\big[\log (1 - D_{\bphi}(G_{\btheta}(\bepsilon)))\big]
	\right],
\end{equation}
where $G_\btheta(\bepsilon)$ is a neural network that converts noise in samples and $D_\bphi(\x)$ classifies samples as fake or not.

 \begin{coupling}[generative adversarial networks]\eod
 Construct a game played between Forger and Curator, with ancillary players Noise and Environment:
  	\begin{itemize}
 		\item \emph{Environment} samples images i.i.d. from $\bP(\x)$.
 		\item \emph{Noise} samples i.i.d. from $\bP(\bepsilon)$.
 		\item \emph{Forger} and \emph{Curator} play parameters $\btheta$ and $\bphi$ respectively.
 		\item \emph{Operator} $G_\btheta(\bepsilon)$ is a neural network that produces fake image $\tilde{\x}=G_\btheta(\bepsilon)$.
 		\item \emph{Operator} $D_\bphi(\tilde{\x})$ is a neural network that estimates the probability that an image is fake.
 		\item The remaining operators compute a loss that \emph{Curator} minimizes and \emph{Forger} maximizes
 		\begin{equation}
 			\vlb(\btheta,\bphi)= \underbrace{\expec_{\x\sim\bP(\x)}\big[\log D_\bphi(\x)\big]}_{\loss_{disc}} 
 			+ \underbrace{\expec_{\bepsilon\sim \bP(\bepsilon)}\Big[\log\big(1-D_\bphi(G_\btheta(\bepsilon))\big)\Big]}_{\loss_{gen}}
 		\end{equation}
 	\end{itemize} 	

 	\vspace{3mm}
	{\small
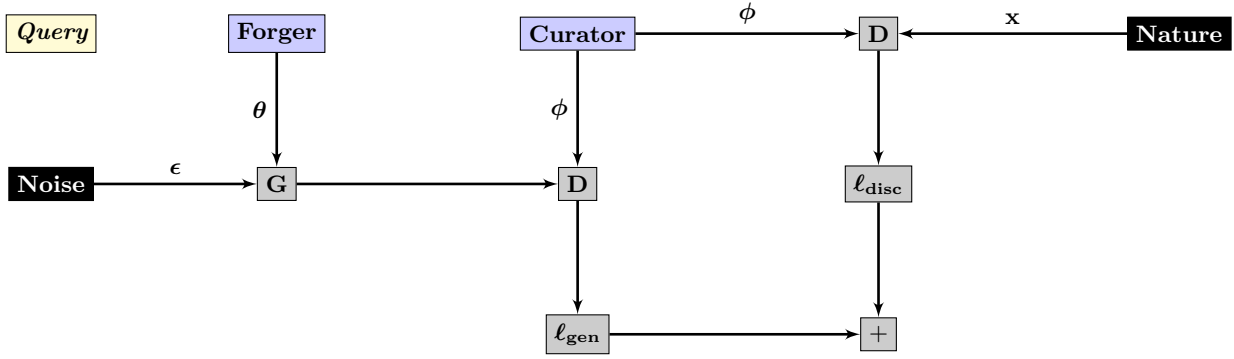
	\begin{tikzpicture}[>=latex']
	    \node[wblock] at (0,0) (glabel) {\textbf{Query}};	    
    	\node[yellow_block] at (0,-2) (player_noise) {\color{white}\rm{\textbf{Noise}}};
	    \node[yellow_block] at (15,0) (player_env)   {\color{white}\rm{\textbf{Nature}}};
	    \node[red_block] at (3,0)  (player1) {\rm{\textbf{Forger}}};
	    \node[red_block] at (7,0)  (player2) {\rm{\textbf{Curator}}};
	    \node[block] at (3,-2)  (block1) {$\mathbf{G}$};
	    \node[block] at (7,-2)  (block2) {$\mathbf{D}$};
	    \node[block] at (11,0) (block4) {$\mathbf{D}$};
	    \node[block] at (11,-2) (block4a) {$\mathbf{\loss_{disc}}$};
	    \node[block] at (7,-4) (block3) {$\mathbf{\loss_{gen}}$};
	    \node[block] at (11,-4) (block5) {$+$};
	    \draw[->,line width=1pt] (player_env) edge node[above] {$\x$}   (block4);
	    \draw[->,line width=1pt] (player1)   edge node[left] {$\btheta$} (block1);
	    \draw[->,line width=1pt] (player2)   edge node[left] {$\bphi$} (block2);
	    \draw[->,line width=1pt] (player2)   edge node[above] {$\bphi$} (block4);
	    \draw[->,line width=1pt] (block1) -- (block2);
	    \draw[->,line width=1pt] (block3) -- (block5);
	    \draw[->,line width=1pt] (block4) -- (block4a);
	    \draw[->,line width=1pt] (block4a) -- (block5);
	    \path[->,line width=1pt] (player_noise) edge node[above] {$\bepsilon$} (block1);
	    \path[->,line width=1pt] (block2) edge node[above] {$\,$} (block3);    
	\end{tikzpicture}
	}

	Note there are two copies of Operator $D$ in the Query graph.
 	The response graph implements the chain rule, with a tweak that multiplies the gradient communicated to \emph{Forger} by $(-1)$ to ensure that \emph{Forger} maximizes the loss that \emph{Curator} is minimizing.\\

	{\small
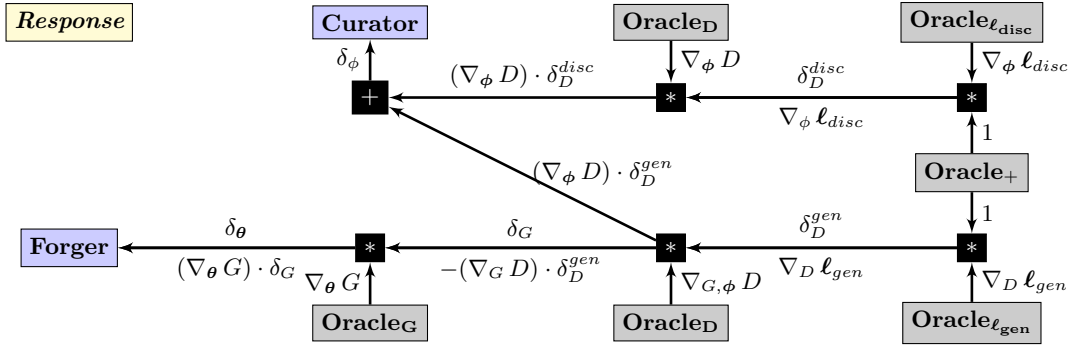
	\begin{tikzpicture}[>=latex']
	    \node[wblock] at (0,0) (glabel) {\textbf{Response}};	    
    	\node[red_block] at (0,-3)  (player1) {\rm{\textbf{Forger}}};
	    \node[red_block] at (4,0)  (player2)  {\rm{\textbf{Curator}}};
	    \node[green_block] at (4,-3)  (block1)  {$\color{white}\mathbf{*}$};
	    \node[green_block] at (8,-3)  (block2)  {$\color{white}\mathbf{*}$};
	    \node[green_block] at (12,-3)  (block3) {$\color{white}\mathbf{*}$};
	    \node[green_block] at (12,-1)  (block4)  {$\color{white}\mathbf{*}$};
	    \node[green_block] at (8,-1)  (block4a)  {$\color{white}\mathbf{*}$};
	    \node[green_block] at (4,-1)  (block5)   {$\color{white}\mathbf{+}$};
	    \node[o_block] at (12,-4) (oracle3) {\rm{\textbf{Oracle$\mathbf{_{\loss_{gen}}}$}}};
	    \node[o_block] at (12,0) (oracle4)  {\rm{\textbf{Oracle$\mathbf{_{\loss_{disc}}}$}}};
	    \node[o_block] at (12,-2) (oracle5) {\rm{\textbf{Oracle$\mathbf{_{+}}$}}};
	    \node[o_block] at (4,-4) (oracle1)  {\rm{\textbf{Oracle$\mathbf{_G}$}}};
	    \node[o_block] at (8,-4) (oracle2)  {\rm{\textbf{Oracle$\mathbf{_D}$}}};
	    \node[o_block] at (8,0) (oracle2a)  {\rm{\textbf{Oracle$\mathbf{_D}$}}};
	
	    \draw[->,line width=1pt] (block1)  edge node[below] {$(\grad_\btheta G)\cdot \delta_{G}$} (player1);
	    \draw[->,line width=1pt] (block1)  edge node[above] {$\delta_\btheta$} (player1);
	    \draw[->,line width=1pt] (block2)  edge node[right] {$(\grad_\bphi D)\cdot \delta_{D}^{gen}$} (block5);
	    \draw[->,line width=1pt] (oracle4) edge node[right] {$\grad_\bphi\loss_{disc}$} (block4);
	    \draw[->,line width=1pt] (block2) edge node[above] {$\delta_{G}$} (block1);
	    \draw[->,line width=1pt] (block2) edge node[below] {$-(\grad_G D)\cdot \delta_{D}^{gen}$} (block1);
	    \draw[->,line width=1pt] (oracle2) edge node[right] {$\grad_{G,\bphi}D$} (block2);
	    \draw[->,line width=1pt] (oracle2a) edge node[right] {$\grad_{\bphi}D$} (block4a);
	    \draw[->,line width=1pt] (oracle1) edge node[left] {$\grad_\btheta G$} (block1);
	    \path[->,line width=1pt] (oracle3) edge node[right] {$\grad_D \loss_{gen}$} (block3);    
	    \draw[->,line width=1pt] (oracle5)  edge node[right] {$1$} (block3);
	    \draw[->,line width=1pt] (oracle5)  edge node[right] {$1$} (block4);
	    \path[->,line width=1pt] (block4) edge node[above] {$\delta_D^{disc}$} (block4a);    
	    \path[->,line width=1pt] (block4a) edge node[above] {$(\grad_{\bphi}D)\cdot \delta_D^{disc}$} (block5);    
	    \path[->,line width=1pt] (block4) edge node[below] {$\grad_\phi \loss_{disc}$} (block4a);        
	    \path[->,line width=1pt] (block3) edge node[above] {$\delta_D^{gen}$} (block2);    
	    \path[->,line width=1pt] (block3) edge node[below] {$\grad_D \loss_{gen}$} (block2);        
		\path[->,line width=1pt] (block5) edge node[left] {$\delta_\phi$} (player2);        
	\end{tikzpicture}
	}
\end{coupling}
\begin{guarantee}
	For a fixed Forger that produces images with probability $\bP_{\text{Forger}}(\x)$, the optimal Curator would assign 
	\begin{equation}
		\label{eq:opt_curator}
		D^*_{\bP_\text{Forger},\bP_\text{Nature}}(\x) = \frac{\bP_\text{Nature}(\x)}
		{\bP_\text{Nature}(\x) + \bP_\text{Forger}(\x)}
	\end{equation}
	The guarantee has two components:
	\begin{enumerate}
		\item For fixed Forger, the Curator in \eqref{eq:opt_curator} is the global optimum for $\vlb$.
		\item The chain rule ensures the correct gradients are communicated to Curator and Forger.
	\end{enumerate}
	It follows that the network converges to a local optimum where Curator represents \eqref{eq:opt_curator} and Forger represents the ``ideal Forger'' that would best fool Curator.
\end{guarantee}
The generative-adversarial network is the first example where the Response graph does not simply backpropagate gradients: the arrow labeled $\delta_G$ is computed as $-(\grad_GD)\cdot \delta_D$, whereas backpropagation would use $(\grad_GD)\cdot \delta_D$. The minus sign arises due to the adversarial relationship between Forger and Curator -- they do not optimize the same objective.

\subsection{Deviator-Actor-Critic (DAC) model}
\label{sec:pg}

As discussed in section~\ref{sec:valprox}, actor-critic algorithms decompose the reinforcement learning problem into two components: the critic, which learns an approximate value function that predicts the total discounted future reward associated with state-action pairs, and the actor, which searches for a policy that maximizes the value appoximation provided by the critic. When the action-space is continuous, a natural approach is to follow the gradient \cite{sutton:99,deisenroth:11,silver:14}. In \cite{sutton:99}, it was shown how to compute the policy gradient given the true value function. Furthermore, sufficient conditions were provided for an approximate value function learned by the critic to yield an unbiased estimator of the policy gradient. More recently \cite{silver:14} provided analogous results for deterministic policies.

The next example of a grammar is taken from \cite{gradprop:15}, which builds on the above work by introducing a third algorithm, Deviator, that directly estimates the gradient of the value function estimated by Critic. 

\begin{coupling}[DAC model]\label{c:compat}\eod
	Construct a game played by Actor, Critic, Deviator, Noise and Environment:\\

	{\small
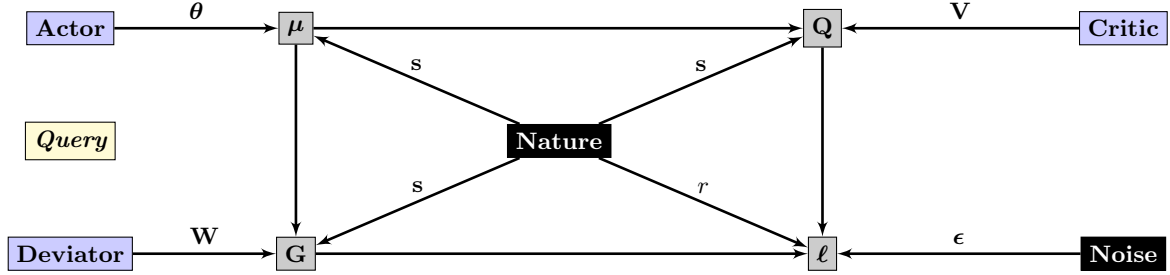
	\begin{tikzpicture}[>=latex']
    	\node[wblock] at (0,-1.5) (glabel) {\textbf{Query}};	    
    	\node[red_block] at (0,0) (actor)     {\rm{\textbf{Actor}}};
    	\node[red_block] at (0,-3) (deviator) {\rm{\textbf{Deviator}}};
    	\node[red_block] at (14,0) (critic)   {\rm{\textbf{Critic}}};
    	\node[yellow_block] at (14,-3) (noise)  {\color{white}\rm{\textbf{Noise}}};
    	\node[yellow_block] at (6.5,-1.5) (env) {\color{white}\rm{\textbf{Nature}}};
    	\node[block] at (3,0)  (block1)   {$\bmu$};
    	\node[block] at (3,-3)  (block2)  {$\Gd$};
    	\node[block] at (10,0)  (block3)  {$\mathbf{Q}$};
    	\node[block] at (10,-3)  (block4) {$\loss$};
    	\draw[->,line width=1pt] (actor) edge node[above] {$\btheta$}  (block1);
    	\draw[->,line width=1pt] (deviator)   edge node[above] {$\Wt$} (block2);    
    	\draw[->,line width=1pt] (critic)   edge node[above] {$\Vt$} (block3);    
    	\draw[->,line width=1pt] (noise)   edge node[above] {$\bepsilon$} (block4);    
    	\draw[->,line width=1pt] (block1)   -- (block2);    
    	\draw[->,line width=1pt] (block1)   -- (block3);    
    	\draw[->,line width=1pt] (block2)   -- (block4);    
    	\draw[->,line width=1pt] (block3)   -- (block4); 
    	\draw[->,line width=1pt] (env)   edge node[above] {$\sta$} (block1);    
    	\draw[->,line width=1pt] (env)   edge node[above] {$\sta$} (block2);    
    	\draw[->,line width=1pt] (env)   edge node[above] {$\sta$} (block3);    
    	\draw[->,line width=1pt] (env)   edge node[above] {$r$} (block4);    
	\end{tikzpicture}
	}

	\begin{itemize}
		\item \emph{Nature} 
		samples states from $\bP(\sta_{t+1}|\sta_t,\act_t)$ and announces rewards $r(\sta_t,\act_t)$ that are a function of the prior state and action; \emph{Noise} samples $\bepsilon\sim N(0,\sigma^2\cdot \bI_d)$.
		\item \emph{Actor}, \emph{Critic} and \emph{Deviator} play parameters $\btheta$, $\Vt$ and $\Wt$ respectively.
		\item \emph{Operator} $\bmu$ is a neural network that computes actions $\act=\bmu_\btheta(\sta)$.
		\item \emph{Operator} $Q^\Vt(\sta,\bmu_\btheta(\sta))$ is a neural network that estimates the value of state-action pairs.
		\item \emph{Operator} $\Gd^\Wt(\sta,\bmu_\btheta(\sta))$ is a neural network that estimates the gradient of the value function.
		\item The remaining \emph{Operator} computes the Bellman gradient error (BGE) which \emph{Critic} and \emph{Deviator} minimize
		\begin{equation}
			\loss_{BGE}(r_t,Q,\tilde{Q},\Gd,\bepsilon)=\left(r_t  + \gamma \tilde{Q}
		- Q - \big\langle \Gd,\bepsilon\big\rangle\right)^2.
		\end{equation}		
	\end{itemize}

	\noindent
	The response graph backpropagates the gradient of $\loss_{BGE}$ to \emph{Critic} and \emph{Deviator}, and communicates the output of \emph{Operator} $\Gd$, which is a \emph{gradient estimate}, to \emph{Actor}:\\

	{\small
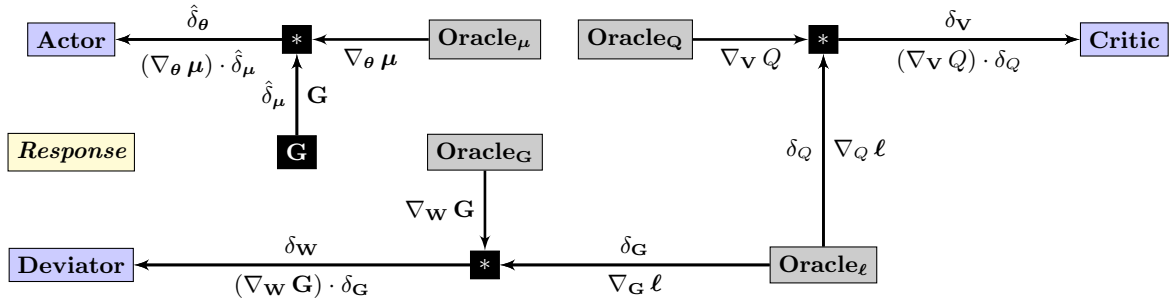
	\begin{tikzpicture}[>=latex']
	    \node[wblock] at (0,-1.5) (glabel) {\textbf{Response}};	    
    	\node[red_block] at (0,0) (actor)      {\rm{\textbf{Actor}}};
	    \node[red_block] at (0,-3) (deviator)  {\rm{\textbf{Deviator}}};
	    \node[red_block] at (14,0) (critic)    {\rm{\textbf{Critic}}};
	    \node[o_block] at (5.5,0)  (oracle1)   {\rm{\textbf{Oracle$_{\bmu}$}}};
	    \node[o_block] at (5.5,-1.5)  (oracle2){\rm{\textbf{Oracle$_\Gd$}}};
	    \node[green_block] at (3,-1.5)  (op2) {$\color{white}\Gd$};
	    \node[o_block] at (7.5,0)  (oracle3) {\rm{\textbf{Oracle$\mathbf{_Q}$}}};
	    \node[o_block] at (10,-3) (oracle4)  {\rm{\textbf{Oracle$_{\loss}$}}};
	    \node[green_block] at (3,0)  (block1) {$\color{white}*$};
	    \node[green_block] at (5.5,-3)  (block2) {$\color{white}*$};    
	    \node[green_block] at (10,0)  (block3) {$\color{white}*$};
	    
	    \draw[->,line width=1pt] (block1)  edge node[below] {$(\grad_\btheta \bmu)\cdot \hat{\delta}_{\bmu}$}  (	actor);
	    \draw[->,line width=1pt] (block1)  edge node[above] {$\hat{\delta}_\btheta$}  (actor);
	    \draw[->,line width=1pt] (block2)  edge node[below] {$(\grad_\Wt \Gd)\cdot \delta_\Gd$} (deviator);    
	    \draw[->,line width=1pt] (block2)  edge node[above] {$\delta_\Wt$} (deviator);    
	    \draw[->,line width=1pt] (block3)  edge node[below] {$(\grad_\Vt Q)\cdot \delta_Q$} (critic);    
	    \draw[->,line width=1pt] (block3)  edge node[above] {$\delta_\Vt$} (critic);    
	    \draw[->,line width=1pt] (oracle1) edge node[below] {$\grad_\btheta\bmu$} (block1);    
	    \draw[->,line width=1pt] (oracle4) edge node[below] {$\grad_\Gd\loss$} (block2);    
	    \draw[->,line width=1pt] (oracle4) edge node[above] {$\delta_\Gd$} (block2);    
	    \draw[->,line width=1pt] (oracle4) edge node[right] {$\grad_Q\loss$} (block3);    
	    \draw[->,line width=1pt] (oracle4) edge node[left] {$\delta_Q$} (block3);    
	    \draw[->,line width=1pt] (oracle2) edge node[left] {$\grad_\Wt\Gd$} (block2);    
	    \draw[->,line width=1pt] (op2) edge node[right] {$\Gd$} (block1);    
	    \draw[->,line width=1pt] (op2) edge node[left] {$\hat{\delta}_{\bmu}$} (block1);    
	    \draw[->,line width=1pt] (oracle3) edge node[below] {$\grad_\Vt Q$} (block3);    
	\end{tikzpicture}
	}
\end{coupling}

Note that instead of backpropagating first-order information in the form of gradient $\grad_{\bmu} \Gd$, the Response graph instead backpropagates zeroth-order information in the form of \emph{gradient-estimate} $\Gd$, which is computed by the Query graph during the feedforward sweep. We therefore write $\hat{\delta}_{\bmu}$ and $\hat{\delta}_{\btheta}$ (instead of $\delta_{\bmu}$ and $\delta_{\btheta})$ to emphasize that the gradients communicated to Actor are estimates. 

As in section~\ref{sec:valprox}, an arrow from Actor to Nature is omitted from the Query graph for simplicity.

\begin{guarantee}
	The guarantee has the following components:
	\begin{enumerate}
		\item \emph{Critic} estimates the value function via TD-learning \cite{sutton:98} with cloning for improved stability \cite{Mnih:2015wq}.
		\item \emph{Deviator} estimates the value gradient via TD-learning and the gradient perturbation trick \cite{gradprop:15}.
		\item \emph{Actor} follows the correct gradient by the policy gradient theorem \cite{sutton:99,silver:14}.		
		\item The internal workings of each neural network are guaranteed correct by the chain rule. 
	\end{enumerate}
	It follows that Critic and Deviator represent the value function and its gradient; and that Actor represents the optimal policy.
\end{guarantee}

Two appealing features of the algorithm are that (i) Actor is insulated from Critic, and only interacts with Deviator and (ii) Critic and Deviator learn different features adapted to representing the value function and its gradient respectively. Previous work used the derivative of the value-function estimate, which is not guaranteed to have compatible function approximation, and can lead to problems when the value-function is estimated using functions such as rectifiers that are not smooth \cite{prokhorov:97,hafner:11,lillicrap:15}.

\subsection{Kickback (truncated backpropagation)}
\label{sec:kb}

Finally we consider Kickback, a biologically-motivated variant of Backprop with reduced communication requirements \cite{bvb:15}. The problem that kickback solves is that backprop requires two distinct kinds of signals to be communicated between units -- feedforward and feedback -- whereas only one signal type -- spikes -- are produced by cortical neurons. Kickback computes an estimate of the backpropagated gradient using the signals generated during the feedforward sweep. Kickback also requires the gradient of the loss with respect to the (one-dimensional) output to be broadcast to all units, which is analogous to the role played by diffuse chemical neuromodulators \cite{schultz:97,pawlak:10,dayan:12}.

\begin{coupling}[kickback]\label{c:pos}\eod
	The query graph is the same as for backpropagation, except that the Operator for each layer produces the additional output $\tau_{i-1}:= \btheta_{i+1}^\intercal \cdot \bbO_{S_{i+1}}$:\\

	{\small
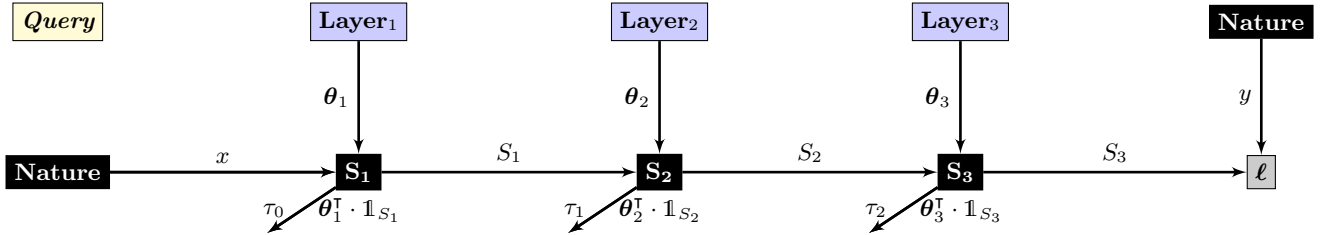
	\begin{tikzpicture}[>=latex']
	    \node[wblock] at (0,0) (glabel) {\textbf{Query}};	    
    	\node[yellow_block] at (0,-2) (player_env1) {\color{white}\rm{\textbf{Nature}}};
	    \node[red_block] at (4,0)  (player1) {\rm{\textbf{Layer$_1$}}};
	    \node[red_block] at (8,0)  (player2) {\rm{\textbf{Layer$_2$}}};
	    \node[red_block] at (12,0) (player3) {\rm{\textbf{Layer$_3$}}};
	    \node[yellow_block] at (16,0) (player_env2) {\color{white}\rm{\textbf{Nature}}};
	    \node[no_block] at (2.5,-3) (out1) {$\,$};
	    \node[no_block] at (6.5,-3) (out2) {$\,$};
	    \node[no_block] at (10.5,-3) (out3) {$\,$};
	    \node[green_block] at (4,-2)  (block1) {$\color{white}\mathbf{S_1}$};
	    \node[green_block] at (8,-2)  (block2) {$\color{white}\mathbf{S_2}$};
	    \node[green_block] at (12,-2) (block3) {$\color{white}\mathbf{S_3}$};
	    \node[block] at (16,-2) (block4) {$\loss$};
	    \draw[->,line width=1pt] (player_env1) -- (block1);
	    \draw[->,line width=1pt] (player1)   edge node[left] {$\btheta_1$} (block1);
	    \draw[->,line width=1pt] (player2)   edge node[left] {$\btheta_2$} (block2);
	    \draw[->,line width=1pt] (player3)   edge node[left] {$\btheta_3$} (block3);
	    \draw[->,line width=1pt] (player_env2)   edge node[left] {$y$}     (block4);
	    \draw[->,line width=1pt] (block1) edge node[above] {$S_1$} (block2);
	    \path[->,line width=1pt] (player_env1) edge node[above] {$x$} (block1);
	    \path[->,line width=1pt] (block2) edge node[above] {$S_2$} (block3);    
	    \path[->,line width=1pt] (block3) edge node[above] {$S_3$} (block4);    
	    \path[->,line width=1pt] (block1) edge node[right] {$\;\btheta_1^\intercal\cdot \bbO_{S_1}$} (out1);    
	    \path[->,line width=1pt] (block2) edge node[right] {$\;\btheta_2^\intercal\cdot \bbO_{S_2}$} (out2);    
	    \path[->,line width=1pt] (block3) edge node[right] {$\;\btheta_3^\intercal\cdot \bbO_{S_3}$} (out3);    
	    \path[->,line width=1pt] (block1) edge node[left] {$\tau_0\;$} (out1);    
	    \path[->,line width=1pt] (block2) edge node[left] {$\tau_1\;$} (out2);    
	    \path[->,line width=1pt] (block3) edge node[left] {$\tau_2\;$} (out3);    
	\end{tikzpicture}
	}

	\begin{itemize}
		\item \emph{Nature} samples labeled data $(x,y)$ from $\bP_{X\times Y}$.
		\item \emph{Layers} by weight matrices $\btheta_i$. The output of the neural network is required to be one-dimensional
		\item \emph{Operators} for each layer compute two outputs: $S_i = \max(0,\btheta_i\cdot S_{i-1})$  \emph{and} $\tau_{i-1}=\btheta_i^\intercal\cdot \bbO_{S_i}$ where $\bbO_a = 1  \text{ if }a\geq 0$ and 0 otherwise.

		\item The task is regression or binary classification with loss given by the mean-squared or logistic error. It follows that the derivative of the loss with respect to the network's output $\beta=\grad_{S_3}\loss$ is a \emph{scalar}.
	\end{itemize}

	\noindent
	The response graph contains a single Oracle that broadcasts the gradient of the loss with respect to the network's output (which is a scalar). Gradient \emph{estimates} for each \emph{Layer} are computed using a mixture of Oracle and local zeroth-order information referred to as \emph{Kicks}:\\

	{\small
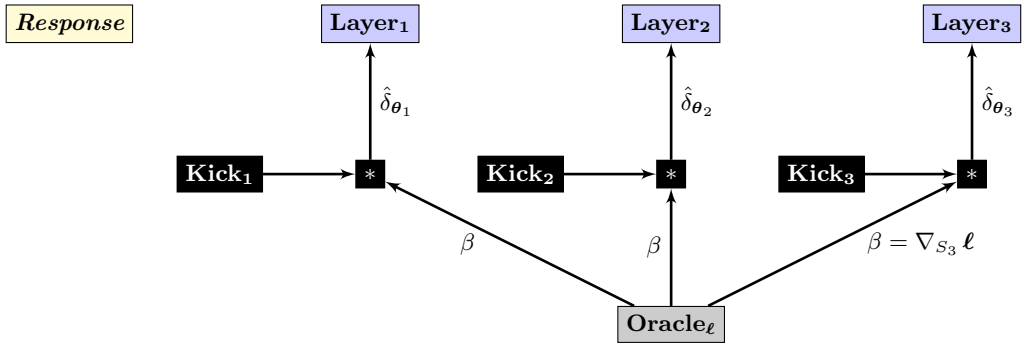
	\begin{tikzpicture}[>=latex']
		\node[wblock] at (2,0) (glabel) {\textbf{Response}};	    
    	\node[green_block] at (4,-2)  (oracle1)  {\color{white}\rm{\textbf{Kick$\mathbf{_1}$}}}; 
	    \node[green_block] at (8,-2)  (oracle2) {\color{white}\rm{\textbf{Kick$\mathbf{_2}$}}};
	    \node[green_block] at (12,-2) (oracle3)  {\color{white}\rm{\textbf{Kick$\mathbf{_3}$}}};
	    \node[o_block] at (10,-4) (oracle_env2) {\rm{\textbf{Oracle$_{\loss}$}}};
	    \node[green_block] at (6,-2)  (block1) {$\color{white}\mathbf{*}$};
	    \node[green_block] at (10,-2)  (block2){$\color{white}\mathbf{*}$};
	    \node[green_block] at (14,-2) (block3) {$\color{white}\mathbf{*}$};
	
	    \node[red_block] at (6,0)  (player1) {\rm{\textbf{Layer$\mathbf{_1}$}}};
	    \node[red_block] at (10,0)  (player2){\rm{\textbf{Layer$\mathbf{_2}$}}};
	    \node[red_block] at (14,0) (player3) {\rm{\textbf{Layer$\mathbf{_3}$}}};

	    \draw[->,line width=1pt] (oracle_env2)   edge node[right] {$\quad\beta=\grad_{S_3}\loss$} (	block3);

	    \draw[->,line width=1pt] (oracle_env2)   edge node[left] {$\beta$} (block2);
	    \draw[->,line width=1pt] (oracle_env2)   edge node[left] {$\beta\quad$} (block1);
	    \draw[->,line width=1pt] (block1)  edge node[right]{$\hat{\delta}_{\btheta_1}$} (player1);
	    \draw[->,line width=1pt] (block2)  edge node[right]{$\hat{\delta}_{\btheta_2}$} (player2);
	    \draw[->,line width=1pt] (block3)  edge node[right]{$\hat{\delta}_{\btheta_3}$} (player3);
	    \draw[->,line width=1pt] (oracle1)  -- (block1);
	    \draw[->,line width=1pt] (oracle2)  -- (block2);
	    \draw[->,line width=1pt] (oracle3)  -- (block3);
	    \end{tikzpicture}
	}\\

	\noindent
	Kick$_i$ is computed using locally available zeroth-order information as follows\\

	{\small
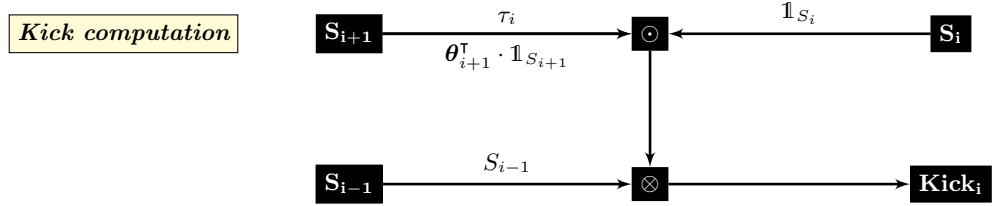
	\begin{tikzpicture}[>=latex']
		\node[wblock] at (0,-2) (glabel) {\textbf{Kick computation}};	
	    \node[green_block] at (3,-4)  (S1)   {\color{white}$\mathbf{S_{i-1}}$};
	    \node[green_block] at (11,-2)  (S2)   {\color{white}$\mathbf{S_i}$};
	    \node[green_block] at (3,-2) (S3)    {\color{white}$\mathbf{S_{i+1}}$};
	    \node[green_block] at (11,-4)  (zero) {\color{white}\rm{\textbf{Kick$\mathbf{_i}$}}};
	    
	    \node[green_block] at (7,-4)  (block1) {$\color{white}\mathbf{\otimes}$};
	    \node[green_block] at (7,-2)  (block2) {$\color{white}\mathbf{\odot}$};
		
	    \draw[->,line width=1pt] (S1) edge node[above]{$S_{i-1}$} (block1);
	    \draw[->,line width=1pt] (S2) edge node[above]{$\bbO_{S_i}$} (block2);
	    \draw[->,line width=1pt] (S3) edge node[above]{$\tau_i$} (block2);
	    \draw[->,line width=1pt] (S3) edge node[below]{$\btheta_{i+1}^\intercal \cdot \bbO_{S_{i+1}}$} (block2);
	    \draw[->,line width=1pt] (block1) -- (zero);
	    \draw[->,line width=1pt] (block2) -- (block1);
	    \end{tikzpicture}
	}\\

	\noindent
	where $\odot$ is coordinatewise multiplication and $\otimes$ is the outer product.
	If $i=1$ then \emph{Nature} is substituted for $S_{i-1}$. If $i=L$ then $S_{i+1}$ is replaced with the scalar value $1$.
\end{coupling}

The loss functions for the layers are not computed in the query graph. Nevertheless, the gradients communicated to the layers by the response graph are exact with respect to the layers' losses, see \cite{bvb:15}. For our purposes it is more convenient to focus on the global objective of the neural network and treat the gradients communicated to the layers as \emph{estimates} of the gradient of the global objective with respect to the layers' weights.

\begin{guarantee}
	Define unit $j$ to be coherent if $\tau_j>0$. A network is coherent if all its units are coherent. A sufficient condition for a rectifier to be coherent is that its weights are positive. 

	The guarantee for Kickback is that, if the network is coherent, then the gradient estimate $\hat{\delta}_{\btheta_i}$ computed using the zeroth-order Kicks has the same sign as the backpropagated error $\delta_{\btheta_i}$ computed using gradients, see \cite{bvb:15} for details. As a result, smalls steps in the direction of the gradient estimates are guaranteed to decrease the network's loss.
\end{guarantee}

\begin{rem}[biological plausibility of kickback]\eod
	Kickback uses a single oracle, analogous to a neuromodulatory signal, in contrast to Backprop which requires an oracle per layer. The rest of the oracles are replaced by kicks -- zeroth-order information from which gradient-estimates are constructed. Importantly, the kick computation for layer $i$ only requires locally available information produced by its neighboring layers $i-1$ and $i+1$ during the feedforward sweep. The feedback signals $\tau_i$ are analogous to the signals transmitted by NMDA synapses.

	Finally, rectifier units with nonnegative weights (for which coherence holds) can be considered a simple model of excitatory neurons \cite{glorot:11,bb:12,balduzzi:14cpm}.
\end{rem}

Two recent alternatives to backprop that also do not rely on backpropagating exact gradients are target propagation \cite{lee:15} and feedback alignment \cite{lillicrap:14}. Target propagation makes do without gradients by implementing autoencoders at each layer. Unfortunately, optimization problems force the authors to introduce a correction term involving \emph{differences} of targets. As a consequence, and in contrast to Kickback, the information required by layers in difference target propagation cannot be computed locally but instead requires recursively backpropagating differences from the output layer. 

Feedback alignment solves a different problem: that feedback and forward weights are required to be equal in backprop (and also in kickback). The authors observe that using random feedback weights can suffice. Unfortunately, as for difference target propagation, feedback alignment still requires separate feedforward and recursively backpropagated training signals, so weight updates are not local.

Unfortunately, at a conceptual level kickback, target propagation and feedback alignment all tackle the wrong problem. The cortex performs reinforcement learning: mammals are not provided with labels, and there is no clearly defined output layer from which signals could backpropagate. A biologically-plausible deep learning algorithm should take advantage of the particularities of the reinforcement learning setting.

\vspace{2mm}
\noindent
\textbf{\sffamily{Acknowledgements.}}
I am grateful to 
Marcus Frean, 
JP Lewis and 
Brian McWilliams 
for useful comments and discussions.

\end{document}